\pdfoutput=1

\documentclass[11pt]{article}

\usepackage[]{naacl2021}

\usepackage{times}
\usepackage{latexsym}

\usepackage[T1]{fontenc}

\usepackage[utf8]{inputenc}

\usepackage{microtype}
\usepackage{times}
\usepackage{latexsym}
\usepackage{graphicx}
\usepackage{amsmath}
\usepackage{amssymb}
\usepackage{booktabs}
\usepackage{multirow}
\usepackage{subfigure}
\usepackage{verbatim}
\usepackage{algorithm}
\usepackage{algorithmic}
\usepackage{makecell}
\usepackage{CJKutf8}
\usepackage{flushend}
\title{DA-Transformer: Distance-aware Transformer}

\author{Chuhan Wu$^\dagger$~~~~Fangzhao Wu$^\ddagger$~~~~\textbf{Yongfeng Huang}$^\dagger$\\
    $^\dagger$Department of Electronic Engineering \& BNRist, Tsinghua University, Beijing 100084, China  \\
     $^\ddagger$Microsoft Research Asia, Beijing 100080, China\\
  \tt{\{wuchuhan15,wufangzhao\}@gmail.com, yfhuang@tsinghua.edu.cn}
  }

\begin{document}

\maketitle

\begin{abstract}
Transformer has achieved great success in the NLP field by composing various advanced models like BERT and GPT.
However, Transformer and its existing variants may not be optimal in capturing token distances because the position or distance embeddings used by these methods usually cannot keep the precise information of real distances,  which may not be beneficial for modeling the orders and relations of contexts.
In this paper, we propose DA-Transformer, which is a distance-aware Transformer that can exploit the real distance.
We propose to incorporate the real distances between tokens to re-scale the raw self-attention weights, which are computed by the relevance between attention query and key.
Concretely, in different self-attention heads the relative distance between each pair of tokens is weighted by different learnable parameters, which control the different preferences on long- or short-term information of these heads.
Since the raw weighted real distances may not be optimal for adjusting self-attention weights, we propose a learnable sigmoid function to map them into re-scaled coefficients that have proper ranges.
We first clip the raw self-attention weights via the ReLU function to keep non-negativity and introduce sparsity, and then multiply them with the re-scaled coefficients to encode real distance information into self-attention.
Extensive experiments on five benchmark datasets show that DA-Transformer can effectively improve the performance of many tasks and outperform the vanilla Transformer and its several variants.

\end{abstract}

\section{Introduction}

Transformer~\cite{vaswani2017attention} has achieved huge success in the NLP field in recent years~\cite{kobayashi2020attention}.
It serves as the basic architecture of various state-of-the-art models like BERT~\cite{devlin2019bert} and GPT~\cite{radford2019language}, and boosts the performance of many tasks like text generation~\cite{koncel2019text}, machine translation~\cite{vaswani2017attention}, and reading comprehension~\cite{xu2019bert}.
Thus, the improvement on the Transformer architecture would be beneficial for many NLP-related fields~\cite{wu2020improving}.

A core component of Transformer is multi-head self-attention, which is responsible for modeling the relations between contexts~\cite{yang2019context,guo2019star}.
However, self-attention is position-agnostic since it does not distinguish the orders of inputs.
Thus, in the vanilla Transformer, position encoding is applied to the input to help Transformer  capture position information.
However, in contrast to recurrent
and convolutional neural networks, it is difficult for vanilla Transformers to be aware of the token distances~\cite{shaw2018self}, which are usually important cues for context modeling. 
Thus, several works explored to incorporate token distance information into Transformer.
For example, Shaw et al.~\shortcite{shaw2018self} proposed to combine the embeddings of relative positions with attention key and value in the self-attention network.
They restricted the maximum relative distance to only keep the precise relative position information within a certain distance. 
\citeauthor{yan2019tener}~\shortcite{yan2019tener} proposed a variant of self-attention network for named entity recognition, which incorporates sinusoidal embeddings of relative position to compute attention weights in a direction- and distance-aware way.
However, the distance or relative position embeddings used by these methods usually cannot keep the precise  information of the real distance, which may not be beneficial for the Transformer to capture word orders and the context relations.


In this paper, we propose a  \underline{\textbf{d}}istance-\underline{\textbf{a}}ware Transformer (DA-Transformer), which can explicitly exploit real token distance information to enhance context modeling by leveraging the relative distances between different tokens to re-scale the raw attention weights before softmax normalization.
More specifically, since global and local context modeling usually have different distance preferences, we propose to learn a different parameter in different attention heads to weight the token distances, which control the preferences of attention heads on long or short distances.
In addition, since the weighted distances may not have been restricted to a proper range, we propose a learnable sigmoid function to map the weighted distances into re-scaled coefficients.
They are further multiplied with the raw attention weights that are clipped by the ReLU function for keeping the non-negativity and introducing sparsity. 
We conduct extensive experiments on five benchmark datasets for different tasks, and the results demonstrate that our approach can effectively enhance the performance of Transformer and outperform its several variants with distance modeling.

The main contributions of this paper include: 
\begin{itemize}
    \item We propose a distance-aware Transformer that uses the real token distances to keep precise distance information in adjusting attention weights for accurate context modeling.
    \item We propose to use different parameters to weight real distances in different attention heads to control their diverse preferences on short-term or long-term information.
    \item We propose a learnable sigmoid function to map the weighted distances into re-scaled coefficients with proper ranges for better adjusting the attention weights.
    \item We conduct extensive experiments on five benchmark datasets and the results validate the effectiveness of our proposed method.
\end{itemize}

\section{Related Work}\label{sec:RelatedWork}
 
 \subsection{Transformer}
To make this paper self-contained, we first briefly introduce the architecture of Transformer, which was initially introduced to the machine translation task~\cite{vaswani2017attention}.
It has become an important basic neural architecture of various state-of-the-art NLP models like BERT~\cite{devlin2019bert} and GPT~\cite{radford2019language}.
The core component of Transformer is multi-head self-attention.
It has $h$ attention heads, where the parameters in each head are independent.
For the $i$-th attention head, it takes a matrix $\mathbf{H}$ as the input.
It first uses three independent parameter matrices $\mathbf{W}_Q^{(i)}$, $\mathbf{W}_K^{(i)}$, and $\mathbf{W}_V^{(i)}$ to respectively transform the input matrix $H$ into the input query $\mathbf{Q}^{(i)}$, key $\mathbf{K}^{(i)}$ and value $\mathbf{V}^{(i)}$, which is formulated as follows:
\begin{equation}
    \mathbf{Q}^{(i)}, \mathbf{K}^{(i)}, \mathbf{V}^{(i)}=\mathbf{H}\mathbf{W}_Q^{(i)}, \mathbf{H}\mathbf{W}_K^{(i)}, \mathbf{H}\mathbf{W}_V^{(i)}.
\end{equation}
Then, it uses a scaled dot-product attention head to process its query, key and value, which is formulated as follows:
\begin{equation}\small
     \mathrm{Attention}(\mathbf{Q}^{(i)}, \mathbf{K}^{(i)}, \mathbf{V}^{(i)})=\mathrm{softmax}(\frac{\mathbf{Q}^{(i)}\mathbf{K}^{(i)\top}}{\sqrt{d}})\mathbf{V}^{(i)},
\end{equation}
where $d$ is the dimension of the vectors in the query and key.
The outputs of the $h$ attention heads are concatenated together and the final output is a linear projection of the concatenated representations, which is formulated as follows:
\begin{equation}\small
\begin{aligned}
     \mathrm{Multihead}(\mathbf{Q}, \mathbf{K}, \mathbf{V})&= \mathrm{Concat(head_1, ..., head_h)}\mathbf{W}_O,\\
    \mathrm{where~~head_i}&= \mathrm{Attention}(\mathbf{Q}^{(i)}, \mathbf{K}^{(i)}, \mathbf{V}^{(i)}),
    \end{aligned}
\end{equation}
where $\mathbf{W}_O$ is an output projection matrix.
In the standard Transformer, a position-wise feed-forward neural network is further applied to the output of multi-head self-attention network.
Its function is formulated as
follows:
\begin{equation}
FFN(\mathbf{x}) = max(0, \mathbf{x}\mathbf{W}_1 + \mathbf{b}_1)\mathbf{W}_2 + \mathbf{b}_2, 
\end{equation}
where $\mathbf{W}_1$, $\mathbf{W}_2$, $\mathbf{b}_1$, $\mathbf{b}_2$ are kernel and bias parameters.
Transformer also employs layer normalization~\cite{ba2016layer} and residual connection~\cite{he2016deep} techniques after the multi-head self-attention and feed-forward neural networks, which are also kept in our method.

Since self-attention network does not distinguish the order and position of input tokens, Transformer adds the sinusoidal embeddings of positions to the input embeddings to capture position information.
However, position embeddings may not be optimal for distance modeling in Transformer because distances cannot be precisely recovered from the dot-product between two position embeddings.

\subsection{Distance-aware Transformer}

Instead of directly using the sinusoidal position embedding~\cite{vaswani2017attention} or the absolute position embedding~\cite{devlin2019bert}, several variants of the Transformer explore to use the relative positions to better model the distance between contexts~\cite{shaw2018self,wang2019self,dai2019transformer,yan2019tener}.
For example, Shaw et al.~\shortcite{shaw2018self} proposed to add the embeddings of relative positions to the attention key and value to capture the relative distance between two tokens.
They only kept the precise distance within a certain range by using a threshold to clip the maximum distance to help generalize to long sequences.
\citeauthor{dai2019transformer}~\shortcite{dai2019transformer} proposed Transformer-XL, which uses another form of relative positional encodings that integrate content-dependent positional scores and a global positional score into the attention weights.
\citeauthor{yan2019tener}~\shortcite{yan2019tener} proposed direction-aware sinusoidal relative position embeddings and used them in a similar way with Transformer-XL.
In addition, they proposed to use the un-scaled attention to better fit the NER task.
However, relative position embeddings may not be optimal for modeling distance information because they usually cannot keep the precise information of real token distances.
Different from these methods, we propose to directly re-scale the attention weights based on the mapped relative distances instead of using sinusoidal position embeddings, which can explicitly encode real distance information to achieve more accurate distance modeling.

\section{DA-Transformer}\label{sec:Model}
In this section, we introduce our proposed \textbf{\underline{d}}istance-\textbf{\underline{a}}ware Transformer (DA-Transformer) approach, which can effectively exploit real token distance information to enhance context modeling. 
It uses a learnable parameter to weight the real distances between tokens in each attention head, and uses a learnable sigmoid function to map the weighted distances into re-scaled coefficients with proper ranges, which are further used to adjust the raw attention weights before softmax normalization. 
The details of DA-Transformer are introduced in the following sections.

\subsection{Head-wise Distance Weighting}

Similar with the standard Transformer, the input of our model is also a matrix that contains the representation of each token, which is denoted as $\mathbf{H}=[\mathbf{h}_1, \mathbf{h}_2, ..., \mathbf{h}_N]$, where $N$ is the length of the sequence.
We denote the real relative distance between the $i$-th and $j$-th positions as $R_{i,j}$, which is computed by $R_{i,j}=|i-j|$.
We can then obtain the relative distance matrix $\mathbf{R}\in \mathbb{R}^{N\times N}$ that describes the relative distance between each pair of positions. 
In each attention head, we use a learnable parameter $w_i$ to weight the relative distance by $\mathbf{R}^{(i)}=w_i\mathbf{R}$, which will be further used to adjust the self-attention weights.
In our method, we stipulate that a more positive $\mathbf{R}^{(i)}$ will amplify the attention weights more strongly while a more negative $\mathbf{R}^{(i)}$ will diminish them more intensively.
Thus, a positive $w_i$ means that this attention head prefers to capture long-distance information, while a negative $w_i$ means that it focuses more on local contexts.
By learning different values of $w_i$, different attention heads may have different preferences on capturing either short-term or long-term contextual information with different intensity.

\begin{figure}[!t]
	\centering 
	\includegraphics[width=0.45\textwidth]{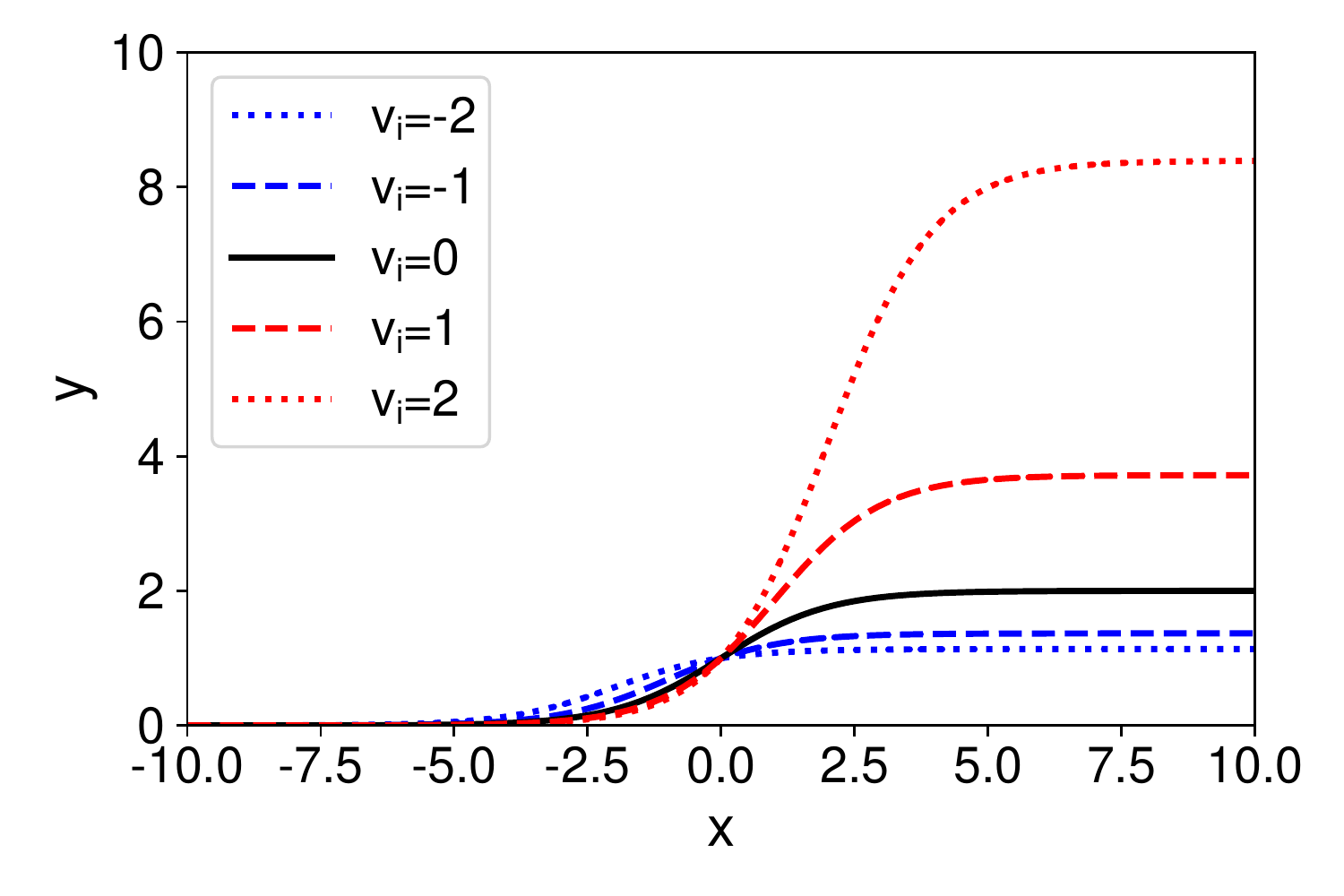} 
\caption{The curves of our learnable sigmoid function under  different $v_i$.}\label{fig.func}
\end{figure}

\subsection{Weighted Distance Mapping}

\begin{figure}[!t]
	\centering 
	\subfigure[$w_i=1, v_i=-1$.]{
	\includegraphics[width=0.22\textwidth]{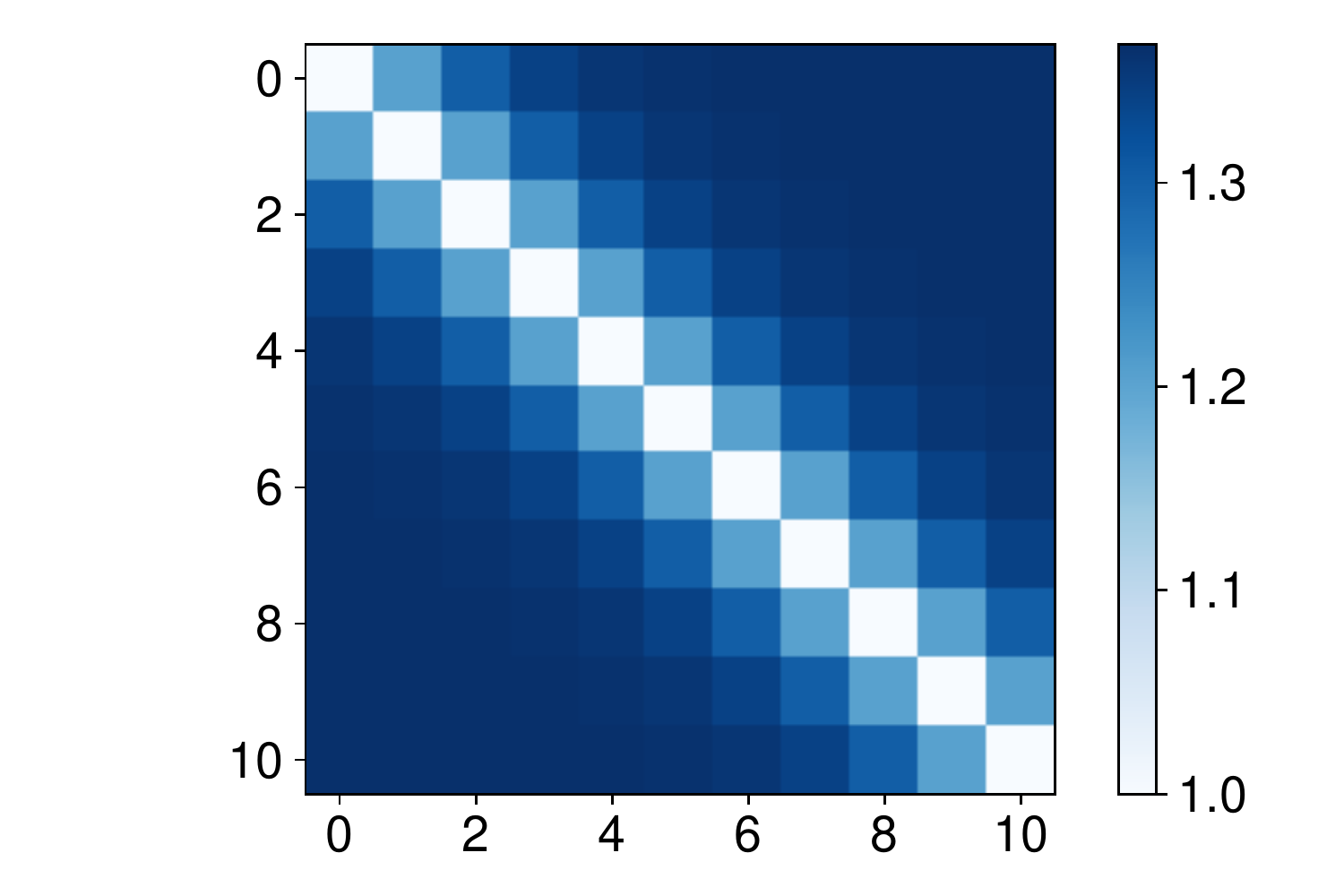} \label{fig.mat1}
	}
		\subfigure[$w_i=-1, v_i=-1$.]{
	\includegraphics[width=0.22\textwidth]{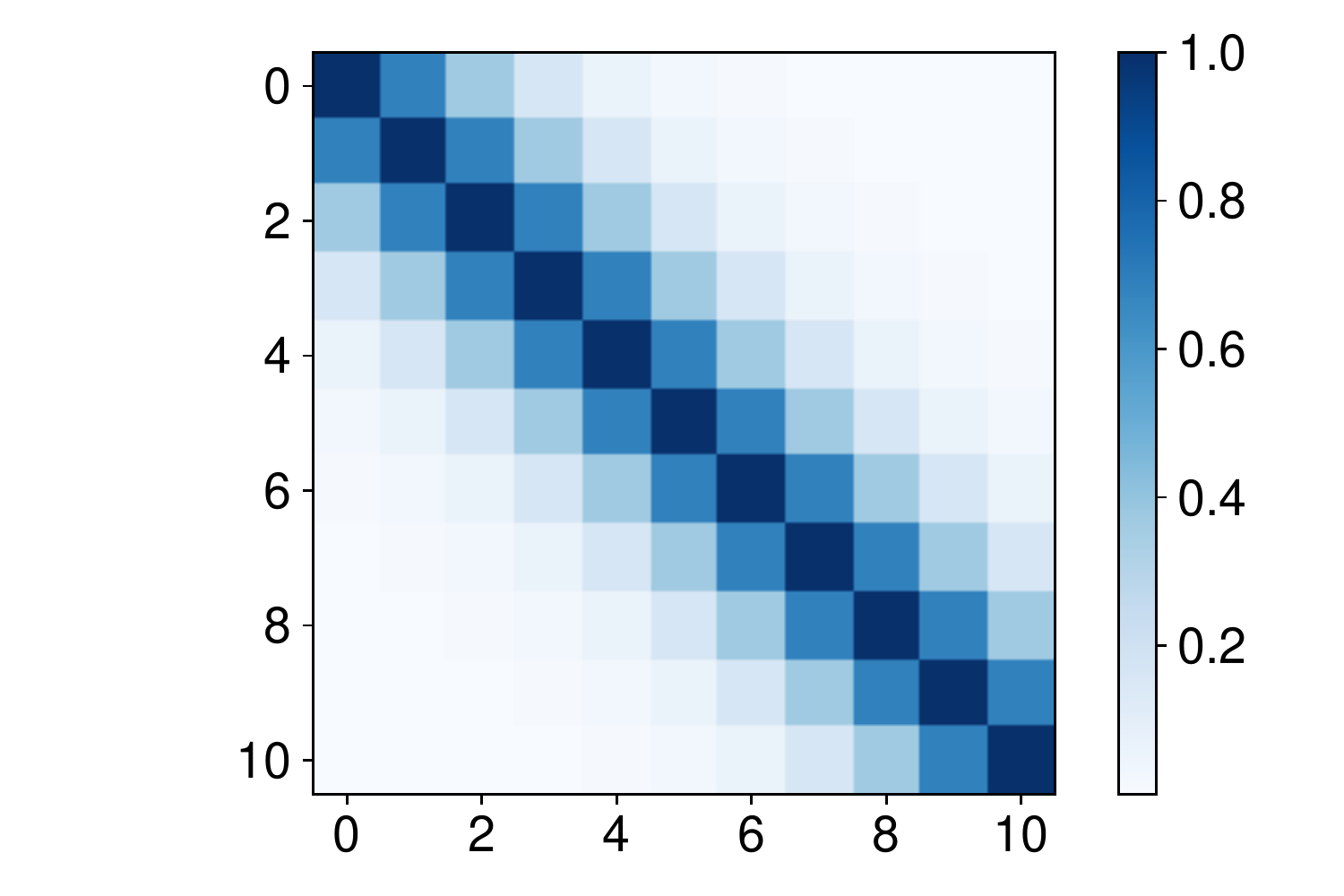} \label{fig.mat2}
	}
		\subfigure[$w_i=1, v_i=1$.]{
	\includegraphics[width=0.22\textwidth]{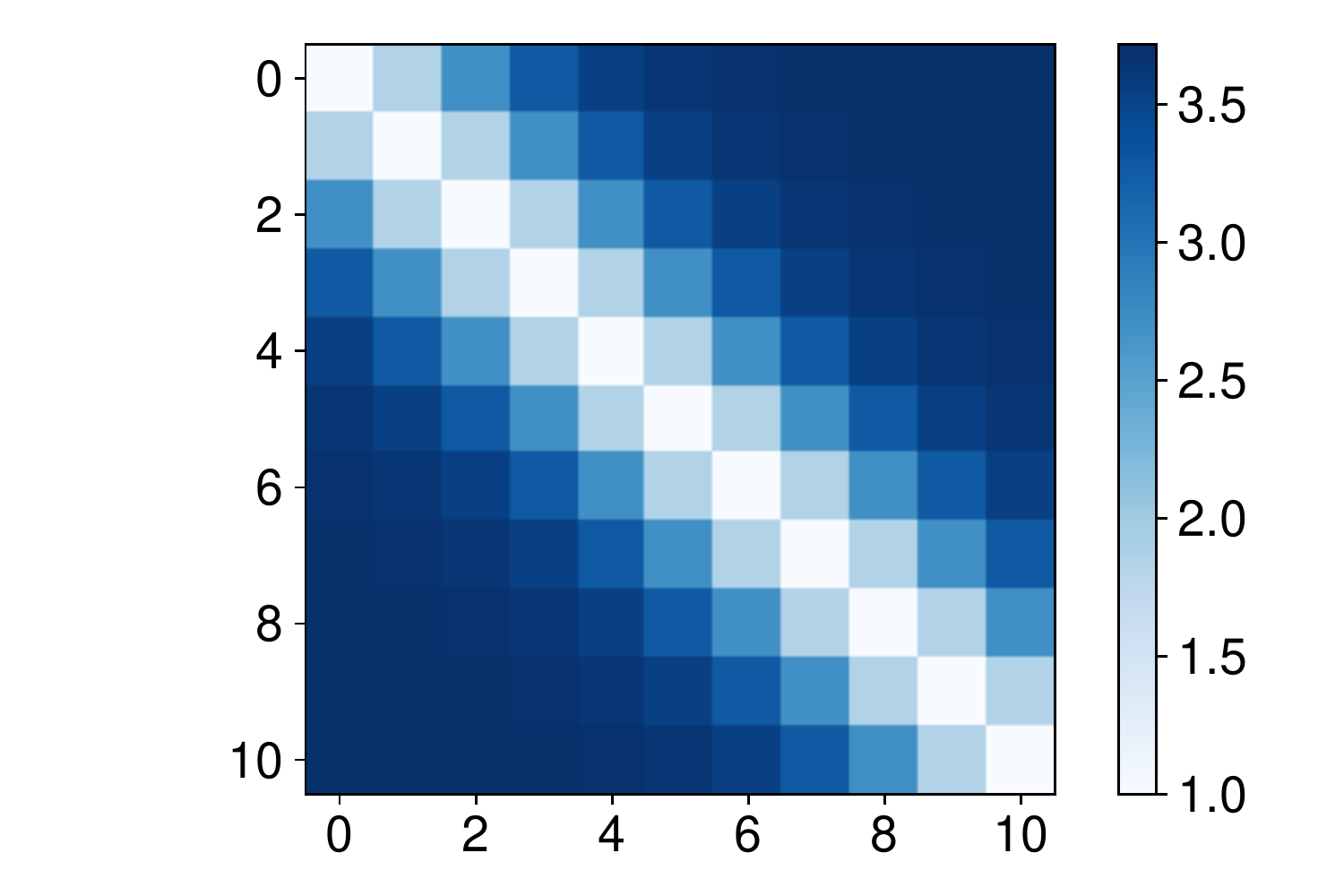} \label{fig.mat3}
	}
		\subfigure[$w_i=-1, v_i=1$.]{
	\includegraphics[width=0.22\textwidth]{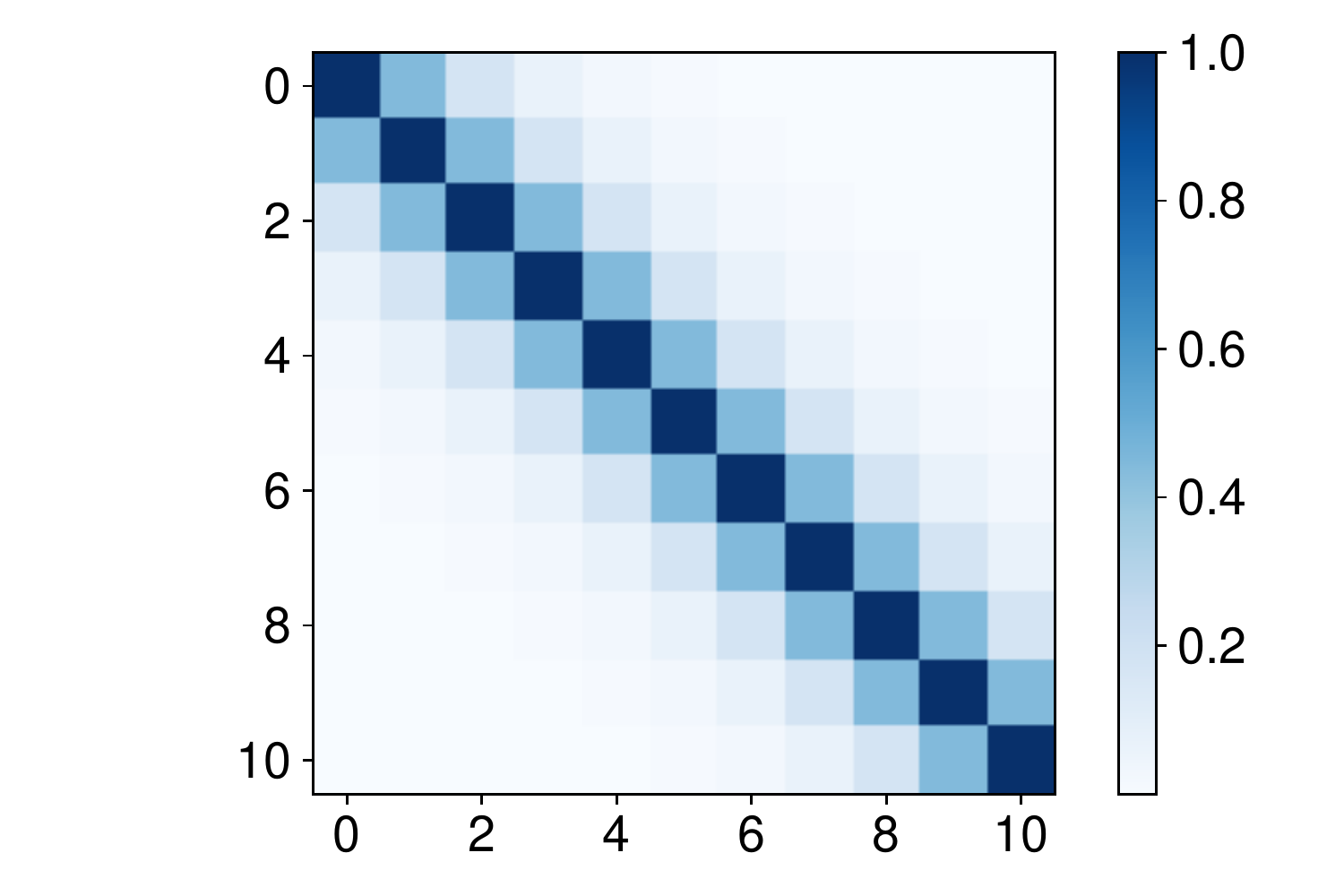} \label{fig.mat4}
	}
\caption{The re-scaled coefficient matrices under different values of $w_i$ and $v_i$. Dark regions indicate that the corresponding attention weights are promoted.}
\end{figure}

Since the raw weighted distances may not be in the proper range for adjusting the attention weights, we need to map them into the re-scaled coefficients via a function $\mathbf{\hat{R}}^{(i)}=f(\mathbf{R}^{(i)})$ that is suitable for adjusting the self-attention weights.
However, it is not a trivial task to design the function $f(\cdot)$ because it needs to satisfy the following requirements:
(1)  $f(0)=1$. 
We stipulate that zero distances do not influence the self-attention weights.
(2) The value of $f(\mathbf{R}^{(i)})$ should be zero when $\mathbf{R}^{(i)}\to-\infty$.
This requirement is to guarantee that if an attention head prefers to capture local information ($w_i<0$), the long-distance information should be surpassed.\footnote{Although the raw negative attention weights may be raised to 0 by $f(\cdot)$, the model can still surpass these attention weights after softmax by increasing the scale of other attention weights.}
(3) The value of $f(\mathbf{R}^{(i)})$ should be limited when $\mathbf{R}^{(i)}\to+\infty$.
This requirement is to ensure that the model is able to process long sequences without over-emphasize distant contexts.
(4) The scale of $f(\cdot)$ needs to be tunable.
This aims to help the model better adjust the intensity of distance information.
(5) The function $f(\cdot)$ needs to be monotone.
To satisfy the five requirements above, we propose a learnable sigmoid function to map the weighted relative distances $\mathbf{R}^{(i)}$, which is formulated as follows:
\begin{equation}\label{eq5}
    f(\mathbf{R}^{(i)}; v_i)=\frac{1+\mathrm{exp}(v_i)}{1+\mathrm{exp}(v_i-\mathbf{R}^{(i)})},
\end{equation}
where $v_i$ is a learnable parameter in this head that controls the upperbound and ascending steepness of this function.
The curves of our learnable sigmoid function under several different values of $v_i$  are plotted in Fig.~\ref{fig.func}.
We can see that the proposed function satisfies all the requirements above. 
In addition, from this figure we find that if $v_i$ is larger, the upperbound of the curve is higher, which means that distance information is more intensive.
When $v_i=0$, it is in fact identical to the standard sigmoid function except for the scaling factor of 2.
By mapping the weighted distances $\mathbf{R}^{(i)}$ via the function $f(\cdot)$, we can obtain the final re-scaled coefficients $\mathbf{\hat{R}}^{(i)}$ in a learnable way.
Several illustrative examples of the re-scaled coefficients under $w_i=\pm 1$ and $v_i=\pm 1$ are respectively shown in Figs.~\ref{fig.mat1}-\ref{fig.mat4}.
We can see that if $w_i$ is positive, long-distance contexts are preferred while short-term contexts are surpassed.
The situation is reversed if $w_i$ turns to negative.
In addition, the coefficients in Fig.~\ref{fig.mat3} have larger dynamic ranges than the coefficients in Fig.~\ref{fig.mat1}, indicating that long-distance information is more dominant in  Fig.~\ref{fig.mat3}.
Moreover, the coefficients in Fig.~\ref{fig.mat4} are ``sharper'' than those in Fig.~\ref{fig.mat2}, which indicates that the model tends to capture shorter distances.

\subsection{Attention Adjustment}

Then, we use the re-scaled coefficients to adjust the raw attention weights that are computed by the dot-product between the query and key, i.e., $\frac{\mathbf{Q}^{(i)}\mathbf{K}^{(i)\top}}{\sqrt{d}}$.
Different from existing methods that add the query-key dot-product with position or distance representations, in our approach we propose to multiply the re-scaled coefficients with the query-key dot-product.
This is because for the tokens whose relations are very weak, if their re-scaled coefficients are large, their final attention weights will be over-amplified if we simply add the re-scaled coefficients to their raw attention weights.
This is not optimal for modeling contextual information because the attention weights of irrelevant contexts cannot be fully surpassed.
However, there are also some problems if we directly multiply the re-scaled coefficients $\mathbf{\hat{R}}^{(i)}$ and the raw attention weights $\frac{\mathbf{Q}^{(i)}\mathbf{K}^{(i)\top}}{\sqrt{d}}$.
This is because the sign of attention weights $\frac{\mathbf{Q}^{(i)}\mathbf{K}^{(i)\top}}{\sqrt{d}}$ is indefinite and the multiplied results cannot accurately reflect the influence of distance information.
Thus, we propose to add a ReLU~\cite{glorot2011deep} activation function to the raw attention weights to keep non-negativity.
In this way, the final output $\mathbf{O}^{(i)}$ of an attention head can be formulated as follows:
\begin{equation}\small \label{eq6}
    \mathbf{O}^{(i)}=\mathrm{softmax}(\frac{\mathrm{ReLU}(\mathbf{Q}^{(i)}\mathbf{K}^{(i)\top})*\mathbf{\hat{R}}^{(i)}}{\sqrt{d}})\mathbf{V}^{(i)},
\end{equation}
where $*$ represents element-wise product. 
The ReLU function can also introduce sparsity to the self-attention because only the  positive attention weights can be amplified by the re-scaled coefficients, which makes the attention weights in our method sharper.
We concatenate the output from the $h$ independent attention heads, and project it into a unified output.
In addition, we keep the same layer normalization and residual connection strategy  as the standard Transformer.

\subsection{Computational Complexity Analysis}

Compared with the standard Transformer, the major additional time  cost is brought by computing the re-scaled coefficients $\mathbf{\hat{R}^{(i)}}$ and using them to adjust the attention weights.
The theoretical time complexity of the two operations in each head is $O(N^2)$, which is much smaller than  the time complexity of computing the attention weights, i.e., $O(N^2\times d)$.
In addition, both Eq. (\ref{eq5}) and Eq. (\ref{eq6}) in our approach can be computed in a vectorized manner.
Thus, the additional time consumption of our method is  very light.
Besides, the increase of parameters is also minimal because we only introduce $2h$ additional parameters, which are usually ignorable compared with the projection matrices like $\mathbf{W}_Q^{(i)}$.
Thus, our approach inherits the efficiency of the Transformer architecture. 

\section{Experiments}\label{sec:Experiments}

\subsection{Datasets and Experimental Settings}
Our experiments are conducted on five benchmark datasets for different tasks.
Four of them are benchmark NLP datasets.
The first one is AG's News\footnote{https://www.di.unipi.it/en/} (denoted as \textit{AG}), which is a news topic classification dataset.
The second one is Amazon Electronics~\cite{he2016ups} (denoted as \textit{Amazon}), which is a dataset for review rating prediction.
The third one is Stanford Sentiment Treebank~\cite{socher2013recursive} (denoted as \textit{SST}).
We use the binary classification version of this dataset.
The fourth one is Stanford Natural Language Inference~\cite{bowman2015large} (\textit{SNLI}) dataset, which is a widely used natural language inference dataset.
The detailed statistics of these datasets are summarized in Table~\ref{table.dataset}.
In addition, we also conduct experiments on a benchmark news recommendation dataset named MIND~\cite{wu2020mind}, aiming to validate the effectiveness of our approach in both text and user modeling.
It contains the news impression logs of 1 million users from Microsoft News\footnote{https://www.msn.com/en-us} from October 12 to November 22, 2019.
The training set contains the logs in the first five weeks except those on the last day which are used for validation.
The rest logs are used for test. 
The key statistics of this dataset are summarized in Table~\ref{table.dataset2}.

\begin{table}[!h]
\centering
	\resizebox{0.48\textwidth}{!}{
\begin{tabular}{lrrrrr}
\hline
\textbf{Dataset}         & \textbf{\# Train} & \textbf{\# Dev.} & \textbf{\# Test} & \textbf{\# Classes} &\textbf{Avg. len.}\\ \hline
 AG             & 108k   & 12k  & 7.6k & 4 & 44 \\
 Amazon      & 40k   & 5k  & 5k  & 5 & 133\\
 SST      & 8k   & 1k   & 2k  & 2 & 19\\
 SNLI      & 55k & 10k & 10k & 2 & 22\\
\hline
\end{tabular}
}
	\caption{Statistics of \textit{AG}, \textit{Amazon}, \textit{SST} and \textit{SNLI} datasets.}\label{table.dataset}

\end{table}

\begin{table}[!h]
\centering
	\resizebox{0.48\textwidth}{!}{
\begin{tabular}{lrlr}
\hline
\textbf{\# Users}            & 1,000,000 & \textbf{Avg. title len.} & 11.52 \\
\textbf{\# News}             & 161,013   & \textbf{\# Click samples}   & 5,597,979 \\
\textbf{\# Impressions}      & 500,000   & \textbf{\# Non-click samples}   & 136,162,621   \\
\hline
\end{tabular}
}
\caption{Statistics of the \textit{MIND} dataset.}\label{table.dataset2}

\end{table}

In our experiments, we use the 300-dimensional Glove~\cite{pennington2014glove} embeddings  for word embedding initialization.\footnote{We do not use contextualized embeddings generated by language models like BERT because we mainly focus on validating the effectiveness of our Transformer architecture.}
The number of attention head is 16, and the output dimension of each attention is 16.
We use one Transformer layer in all experiments.
On the \textit{AG}, \textit{SST} and \textit{SNLI} datasets, we directly apply Transformer-based methods to the sentences.
On the \textit{Amazon} dataset, since reviews are usually long documents, we use Transformers in a hierarchical way by learning sentence representations from words via a word-level Transformer first and then learning document representations from sentences via a sentence-level Transformer.
On the \textit{MIND} dataset, following~\cite{wu2019nrms,wu2020sentirec} we also use a hierarchical model architecture that first learns representations of historical clicked news and candidate news from their titles with a word-level Transformer, then learns user representations from the representations of clicked news with a news-level Transformer, and final matches user and candidate news representations to compute click scores.\footnote{Both the word-level and news-level Transformers contain one self-attention layer.}
We use the same model training strategy with negative sampling techniques as NRMS~\cite{wu2019nrms}.
On all datasets we use Adam~\cite{kingma2014adam} as the optimization algorithm and the learning rate is 1e-3. 
On the \textit{AG}, \textit{Amazon}, \textit{SST} and \textit{SNLI} datasets, accuracy and macro-Fscore are used as the performance metric.
On the \textit{MIND} dataset, following~\cite{wu2019nrms} we use the average AUC, MRR, nDCG@5 and nDCG@10 scores of all sessions as the metrics.
Each experiment is repeated 5 times independently and the average results with standard deviations are reported.

\begin{table*}[!t]
\centering
\begin{tabular}{lcccc}
\hline
\multicolumn{1}{c}{\multirow{2}{*}{\textbf{Methods}}} & \multicolumn{2}{c}{\textbf{AG}} & \multicolumn{2}{c}{\textbf{Amazon}}  \\ \cline{2-5} 
\multicolumn{1}{c}{}                                  & Accuracy        & Macro-F       & Accuracy          & Macro-F             \\ \hline
Transformer                                           & 93.01$\pm$0.13           & 93.00$\pm$0.13         & 65.15$\pm$0.40             & 42.14$\pm$0.41               \\
Transformer-RPR                                       & 93.14$\pm$0.12           & 93.13$\pm$0.13         & 65.29$\pm$0.38             & 42.40$\pm$0.40            \\
Transformer-XL                                        & \underline{93.35}$\pm$0.10           & \underline{93.34}$\pm$0.11         & \underline{65.50}$\pm$0.40             & \underline{42.88}$\pm$0.43                 \\
Adapted Transformer                                                 & 93.28$\pm$0.13           & 93.27$\pm$0.14         & 65.47$\pm$0.39             & 42.69$\pm$0.42          \\ \hline
*DA-Transformer                                      & \textbf{93.72}$\pm$0.11           & \textbf{93.70}$\pm$0.12         & \textbf{66.38}$\pm$0.39             & \textbf{44.29}$\pm$0.40             \\ \hline
\end{tabular}
\caption{Results on \textit{AG} and \textit{Amazon}. *Improvement over the underlined second best results  is significant at $p<0.05$.}
\label{table.result}

\end{table*}

\begin{table*}[!t]
\centering
\begin{tabular}{lcccc}
\hline
\multicolumn{1}{c}{\multirow{2}{*}{\textbf{Methods}}}  & \multicolumn{2}{c}{\textbf{SST}} & \multicolumn{2}{c}{\textbf{SNLI}} \\ \cline{2-5} 
\multicolumn{1}{c}{}                      & Accuracy        & Macro-F        & Accuracy         & Macro-F        \\ \hline
Transformer                                                 & 89.67$\pm$0.22           & 89.59$\pm$0.24          & 81.45$\pm$0.30            & 81.42$\pm$0.31          \\
Transformer-RPR                                          & 89.94$\pm$0.19           & 89.90$\pm$0.20          & 82.20$\pm$0.31            & 82.18$\pm$0.31          \\
Transformer-XL                                    & 90.06$\pm$0.20           & 90.02$\pm$0.21          & \underline{83.19}$\pm$0.29            & \underline{83.15}$\pm$0.30          \\
Adapted Transformer                                 & \underline{90.15}$\pm$0.19           & \underline{90.10}$\pm$0.1          & 82.35$\pm$0.28            & 82.31$\pm$0.30          \\ \hline
*DA-Transformer                               & \textbf{90.49}$\pm$0.17           & \textbf{90.43}$\pm$0.19          & \textbf{84.18}$\pm$0.27            & \textbf{84.16}$\pm$0.29          \\ \hline
\end{tabular}
\caption{Results on \textit{SST} and \textit{SNLI}. *Improvement over the underlined second best results  is significant at $p<0.05$.}
\label{table.result}

\end{table*}

\begin{table*}[!t]
\centering
\begin{tabular}{lcccc}
\hline
\multicolumn{1}{c}{\textbf{Methods}} & AUC                                & MRR                                & \small{nDCG@5}                             & \small{nDCG@10}        \\ \hline
Transformer                          & 67.76$\pm$0.18                              & 33.05$\pm$0.16                              & 35.94$\pm$0.19                              & 41.63$\pm$0.20          \\
Transformer-RPR                      & 67.81$\pm$0.16                              & 33.10$\pm$0.17                              & 35.98$\pm$0.20                              & 41.65$\pm$0.21          \\
Transformer-XL                       & \underline{67.92}$\pm$0.16                              & \underline{33.15}$\pm$0.16                              & \underline{36.04}$\pm$0.20                              & \underline{41.70}$\pm$0.19          \\
Adapted Transformer                                & 67.70$\pm$0.22                              & 33.01$\pm$0.20                              & 35.89$\pm$0.17                              & 41.58$\pm$0.23          \\ \hline
*DA-Transformer                     & \textbf{68.32}$\pm$0.15 & \textbf{33.36}$\pm$0.16 & \textbf{36.34}$\pm$0.14 & \textbf{42.07}$\pm$0.17 \\ \hline
\end{tabular}
 \caption{Results on the \textit{MIND} dataset. *Improvement over the underlined second best results is significant at $p<0.05$.}
\label{table.result2}

\end{table*}

\subsection{Performance Evaluation}

We compare our proposed \textit{DA-Transformer} method with several baseline methods, including:
(1) \textit{Transformer}~\cite{vaswani2017attention}, the vanilla Transformer architecture, where sinusoidal positional embeddings are used.
(2)  \textit{Transformer-RPR}~\cite{shaw2018self}, a variant of Transformer with relative position representations. 
(3)  \textit{Transformer-XL}~\cite{dai2019transformer}, a variant of Transformer that consists of a segment-level recurrence mechanism
and a sinusoidal relative position encoding scheme. 
(4)  \textit{Adapted Transformer}~\cite{yan2019tener}, a variant of Transformer that uses direction- and distance-aware position encoding. 
The results of our approach and these methods on the five datasets are respectively shown in Tables~\ref{table.result} and~\ref{table.result2}.
From the results, we have several observations.

First, compared with the vanilla Transformer, the compared methods that consider distance information consistently achieve better performance.
It shows that distance information is very important in context modeling.
Second, among the methods with distance information, the performance of \textit{Transformer-RPR} is lower than the others.
This may be because \textit{Transformer-RPR} does not keep the precise long-distance information.
Third, by comparing \textit{Transformer-XL} and \textit{Adapted Transformer}, we find that the performance of \textit{Adapted Transformer} is better on the SST dataset, while \textit{Transformer-XL} is better on other datasets.
This is probably because \textit{Adapted Transformer} is more suitable for modeling local contexts and the sentences in the \textit{SST} dataset are usually short, while \textit{Transformer-XL} may be more appropriate for modeling long sequences. 
Fourth, our method  consistently achieves better performance on the five datasets, and its improvement over the second best method is statistically significant (t-test p<0.05).
This is because our method can explicitly encode real distance information rather than using positional encoding, making the modeling of distance more accurate.

\begin{figure}[!t]
	\centering 
	\includegraphics[width=0.4\textwidth]{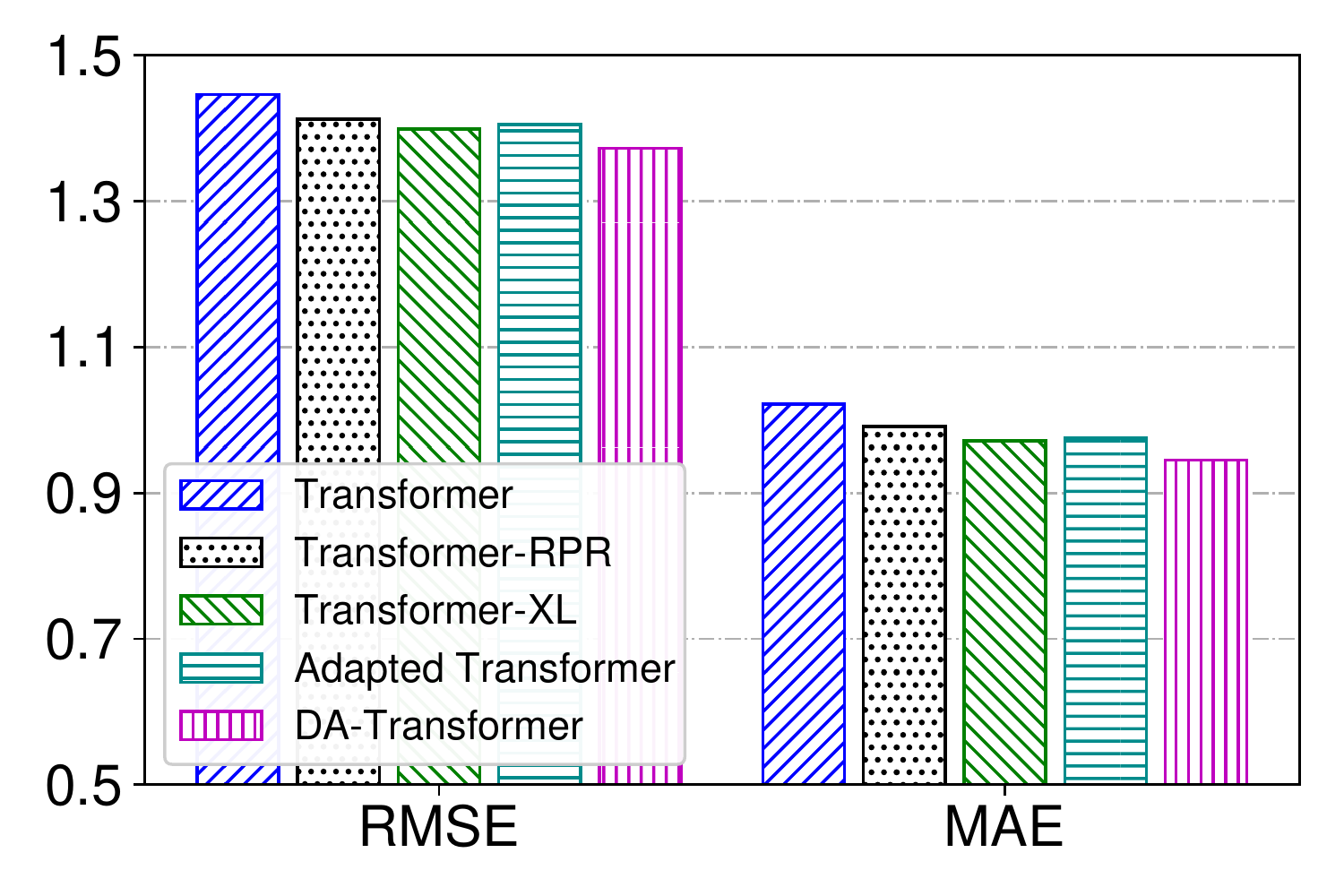} 
	
\caption{Performance comparison of rating regression on \textit{Amazon}. Lower scores indicate better performance.}\label{fig.reg}
\end{figure}

We further compare the performance of different methods in a rating regression task on the \textit{Amazon} dataset.
The results are shown in Fig.~\ref{fig.reg}.
From Fig.~\ref{fig.reg} we observe similar patterns with the results in classification tasks, which  validate the generality of our DA-Transformer in different genres of tasks.

\begin{figure*}[!t]
	\centering 
	\subfigure[\textit{AG}.]{
	\includegraphics[height=1.16in]{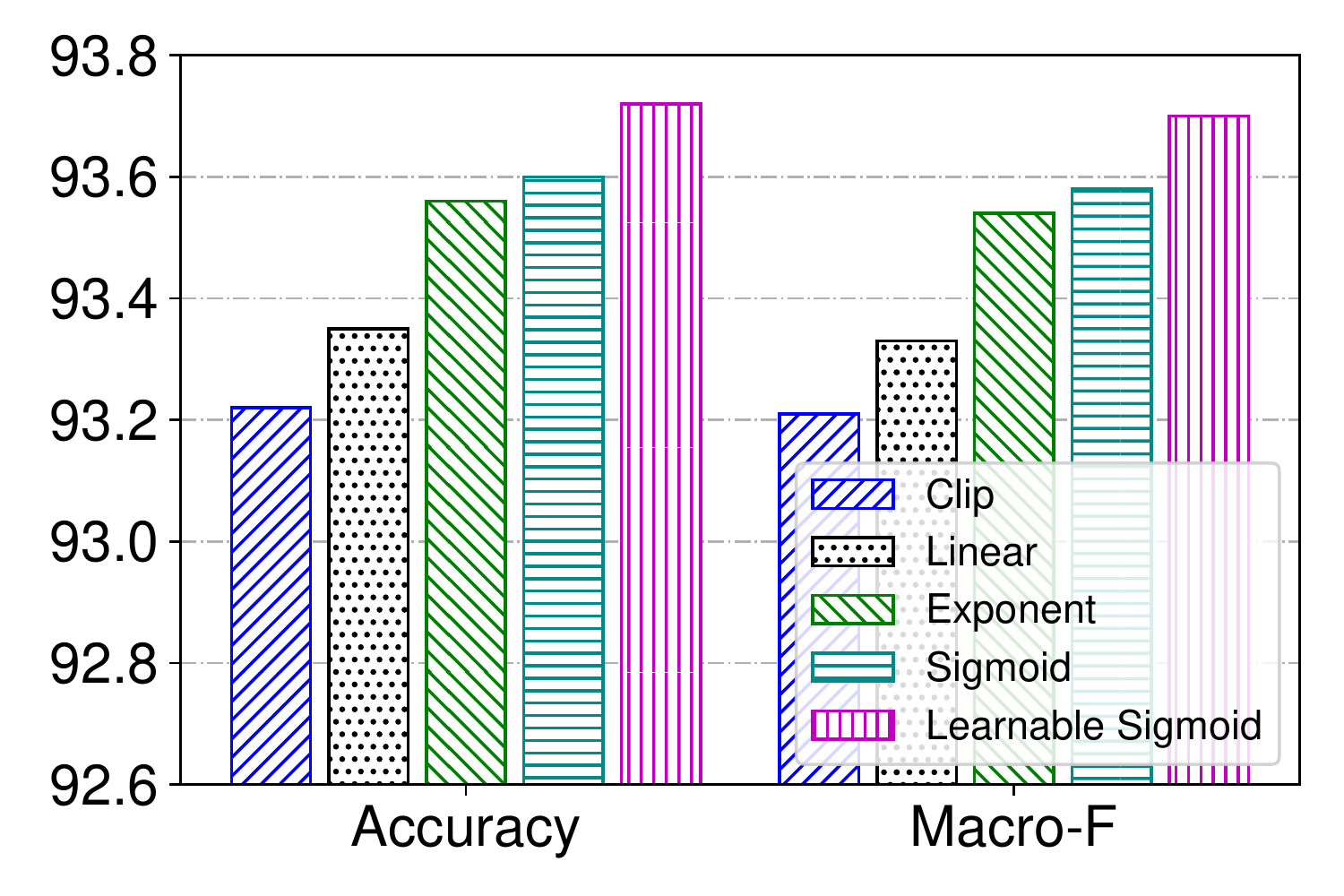} 
	}
		\subfigure[\textit{Amazon}.]{
	\includegraphics[height=1.16in]{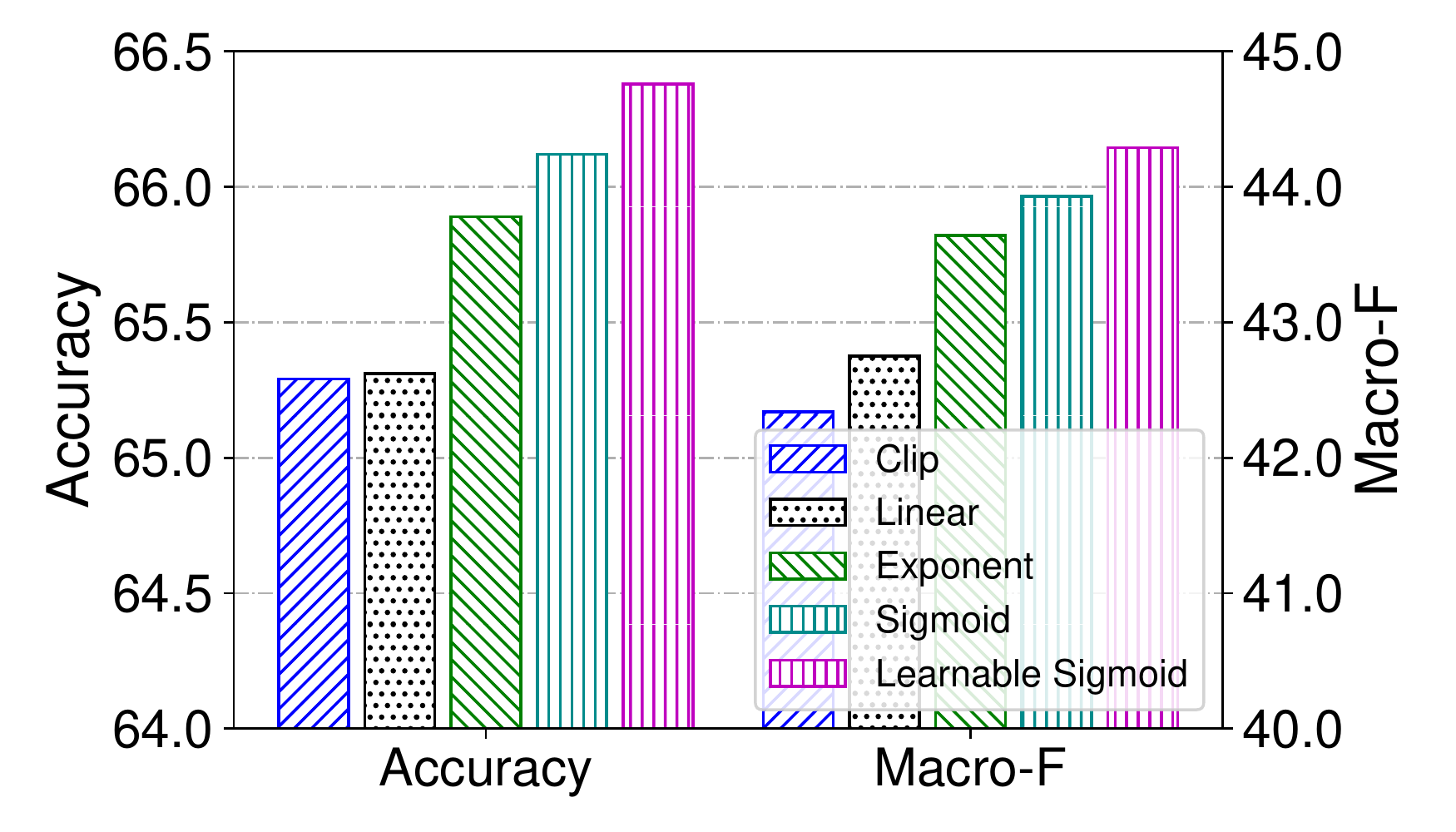} 
	}
		\subfigure[\textit{MIND}.]{
	\includegraphics[height=1.16in]{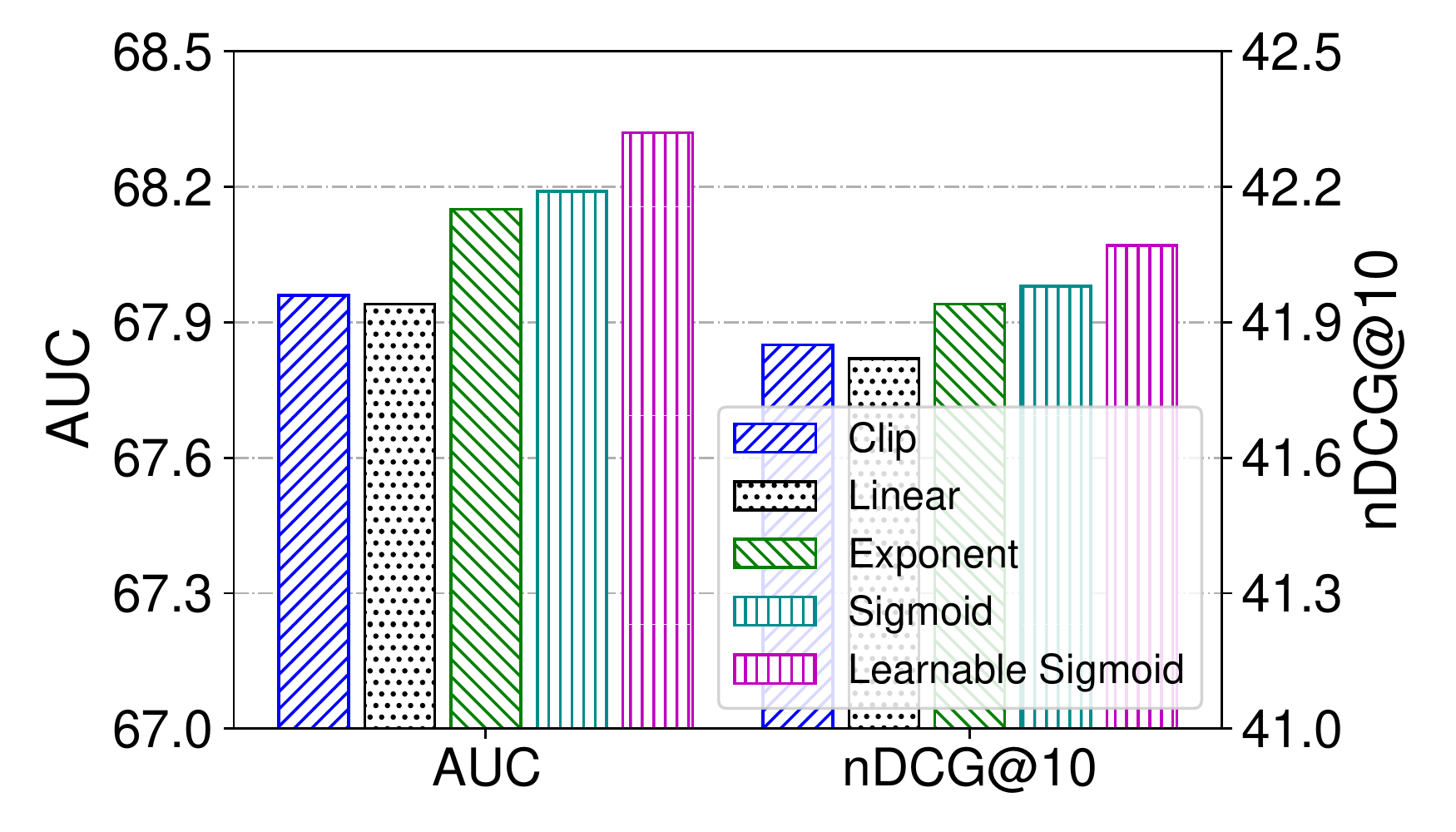} 
	}
\caption{Influence of using different mapping functions.}\label{fig.positionfunc1}
\end{figure*}

\begin{figure*}[!t]
	\centering 
	\subfigure[\textit{AG}.]{
	\includegraphics[height=1.16in]{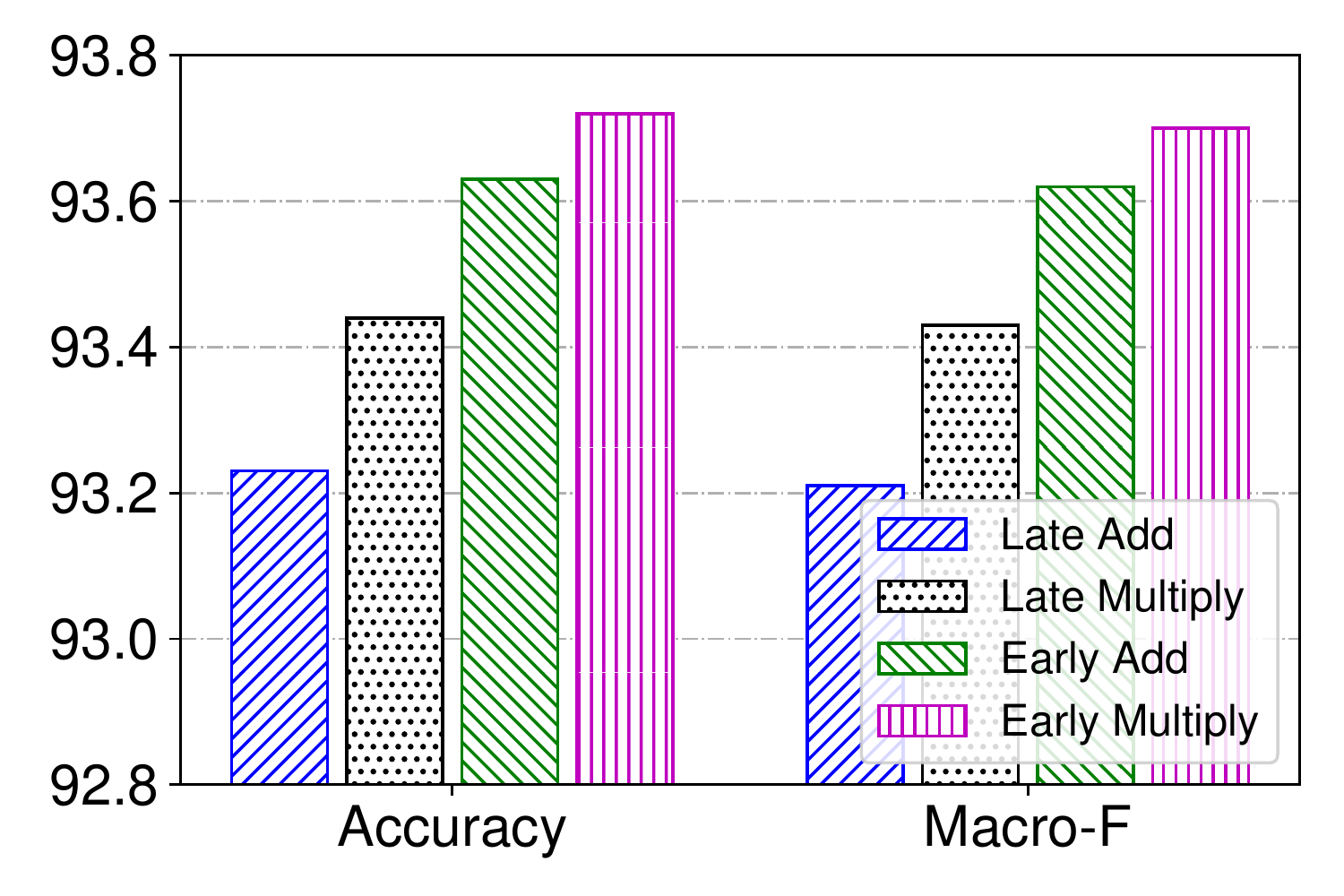} 
	}
		\subfigure[\textit{Amazon}.]{
	\includegraphics[height=1.16in]{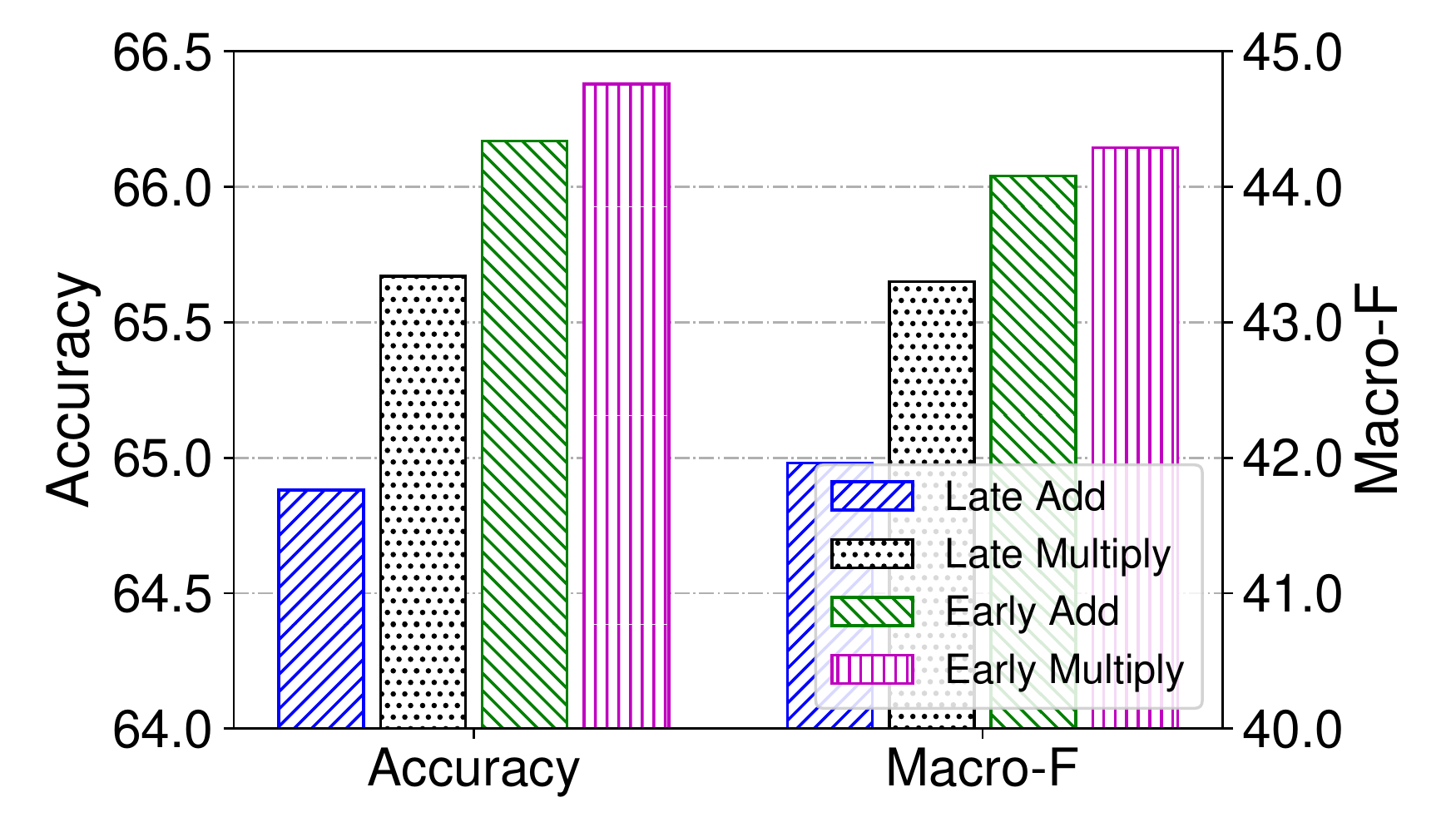} 
	}
		\subfigure[\textit{MIND}.]{
	\includegraphics[height=1.16in]{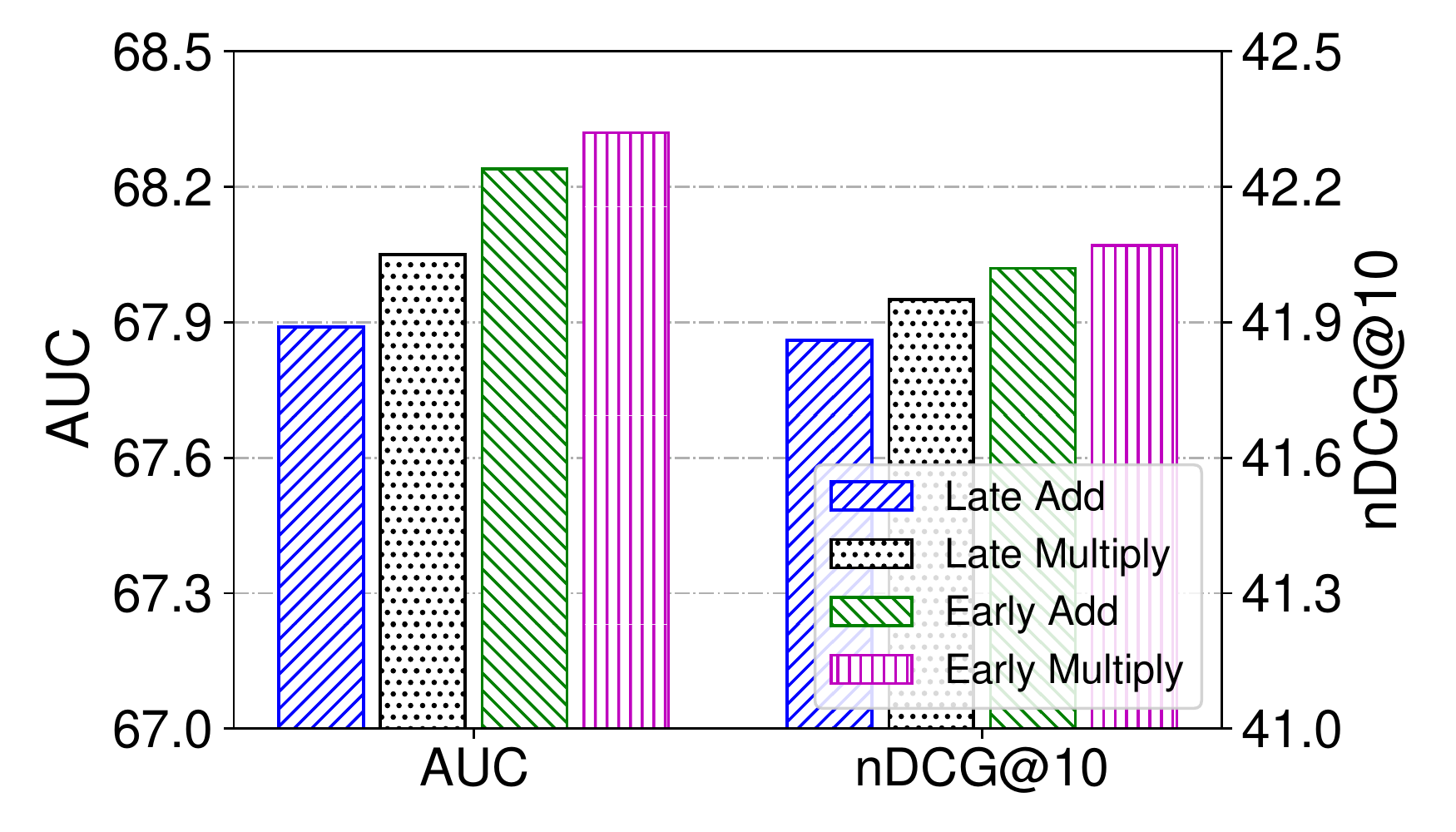} 
	}
\caption{Influence of using different attention adjusting methods.}\label{fig.positionfunc2}
\end{figure*}

\begin{figure}[!t]
	\centering 
	\subfigure[$w_i$.]{
	\includegraphics[width=0.22\textwidth]{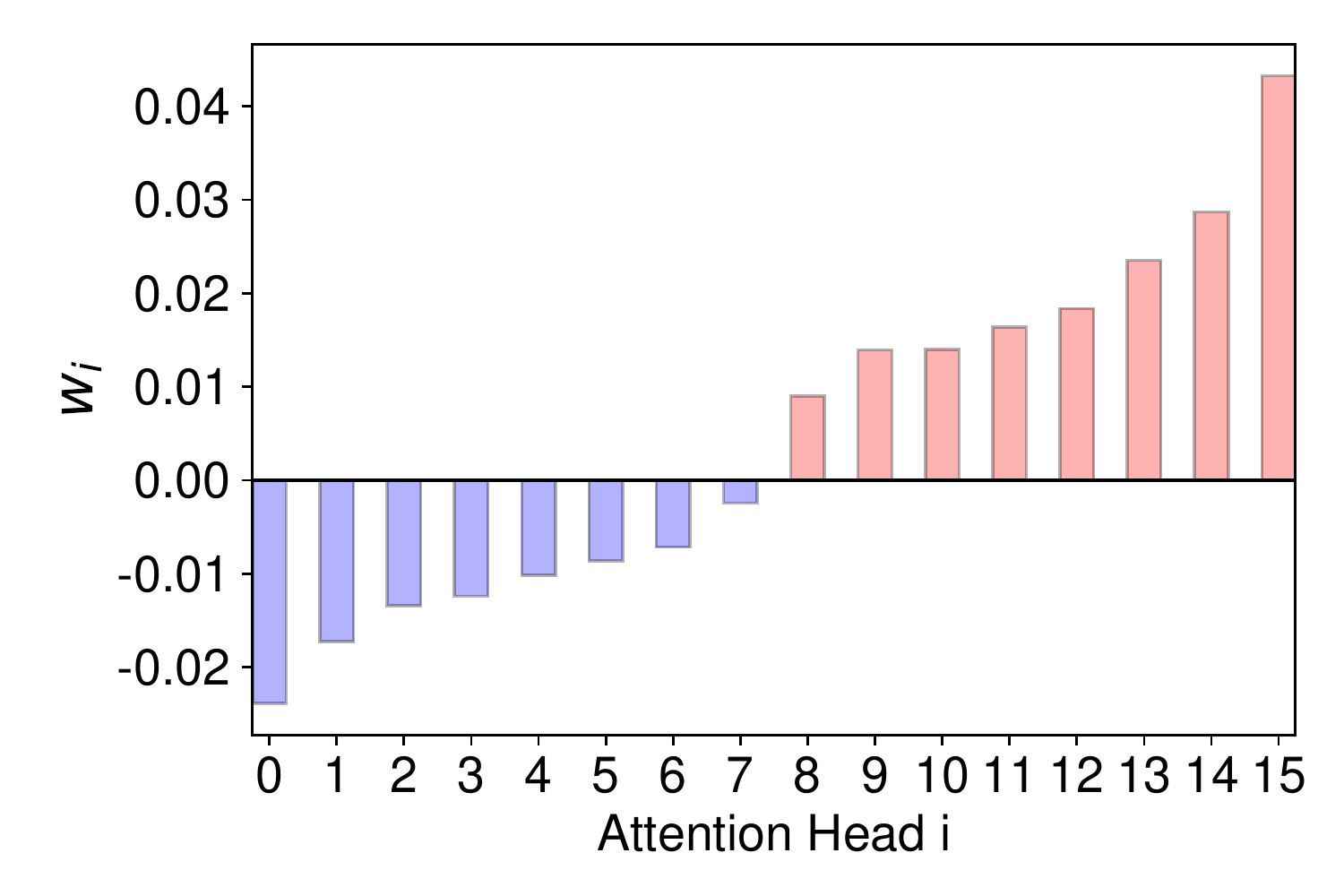} 
	}
		\subfigure[$v_i$.]{
	\includegraphics[width=0.22\textwidth]{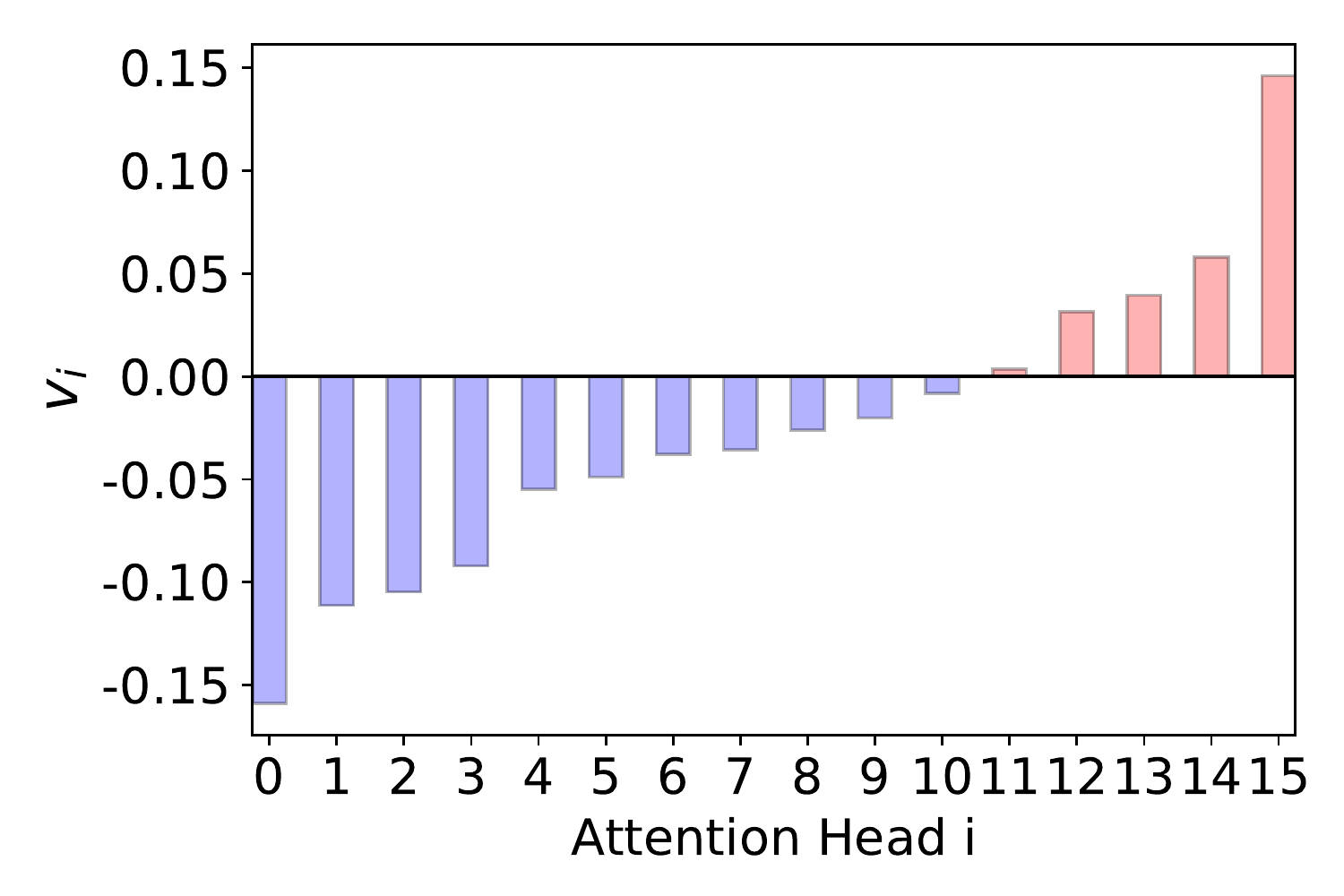} 
	}
\caption{The  weights learned by different attention heads on the \textit{AG} dataset.}\label{fig.aghead}
\end{figure}


\begin{figure}[!t]
	\centering 
	\subfigure[Word-level $w_i$.]{
	\includegraphics[width=0.22\textwidth]{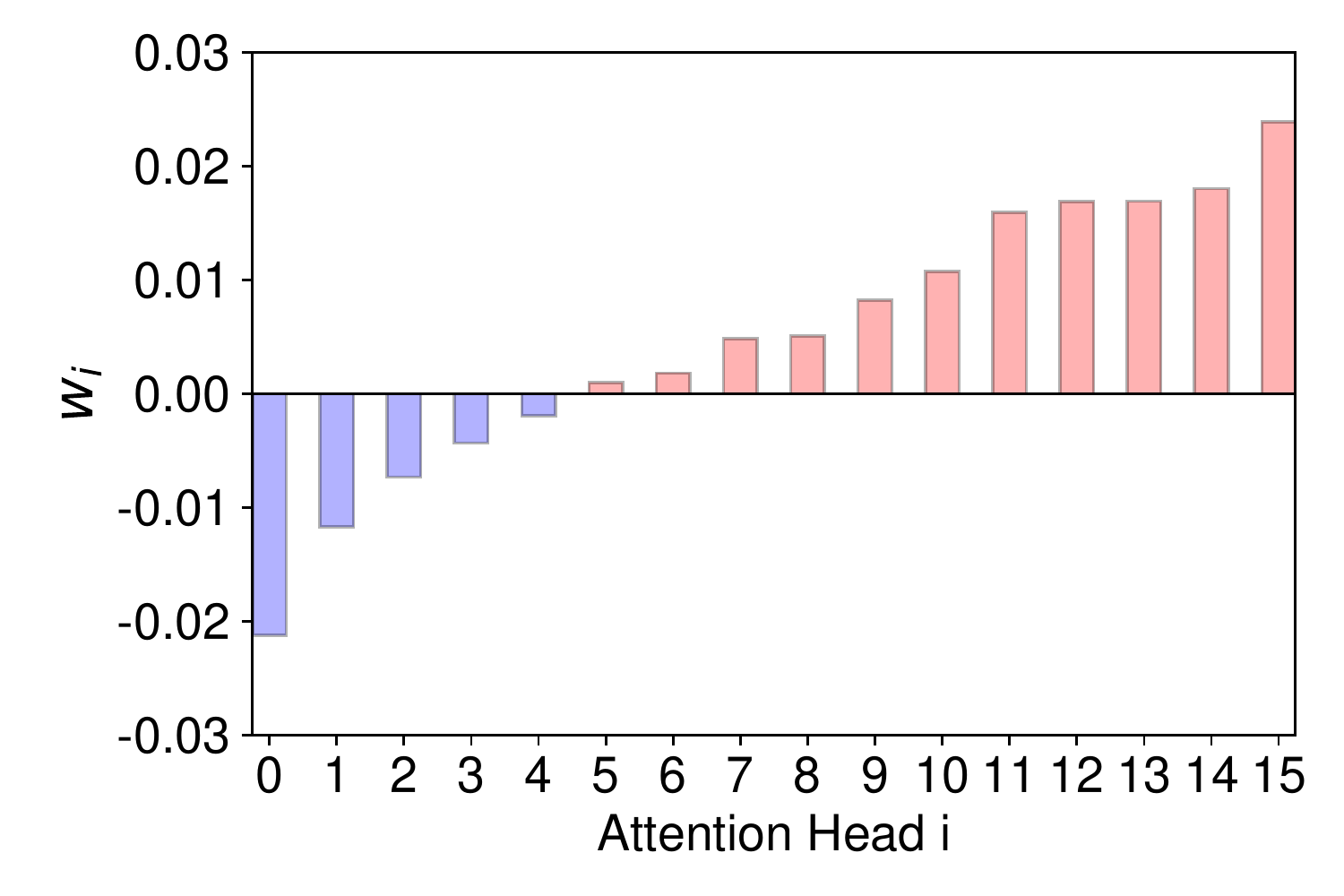} \label{fig.newshead1}
	}
		\subfigure[News-level $w_i$.]{
	\includegraphics[width=0.22\textwidth]{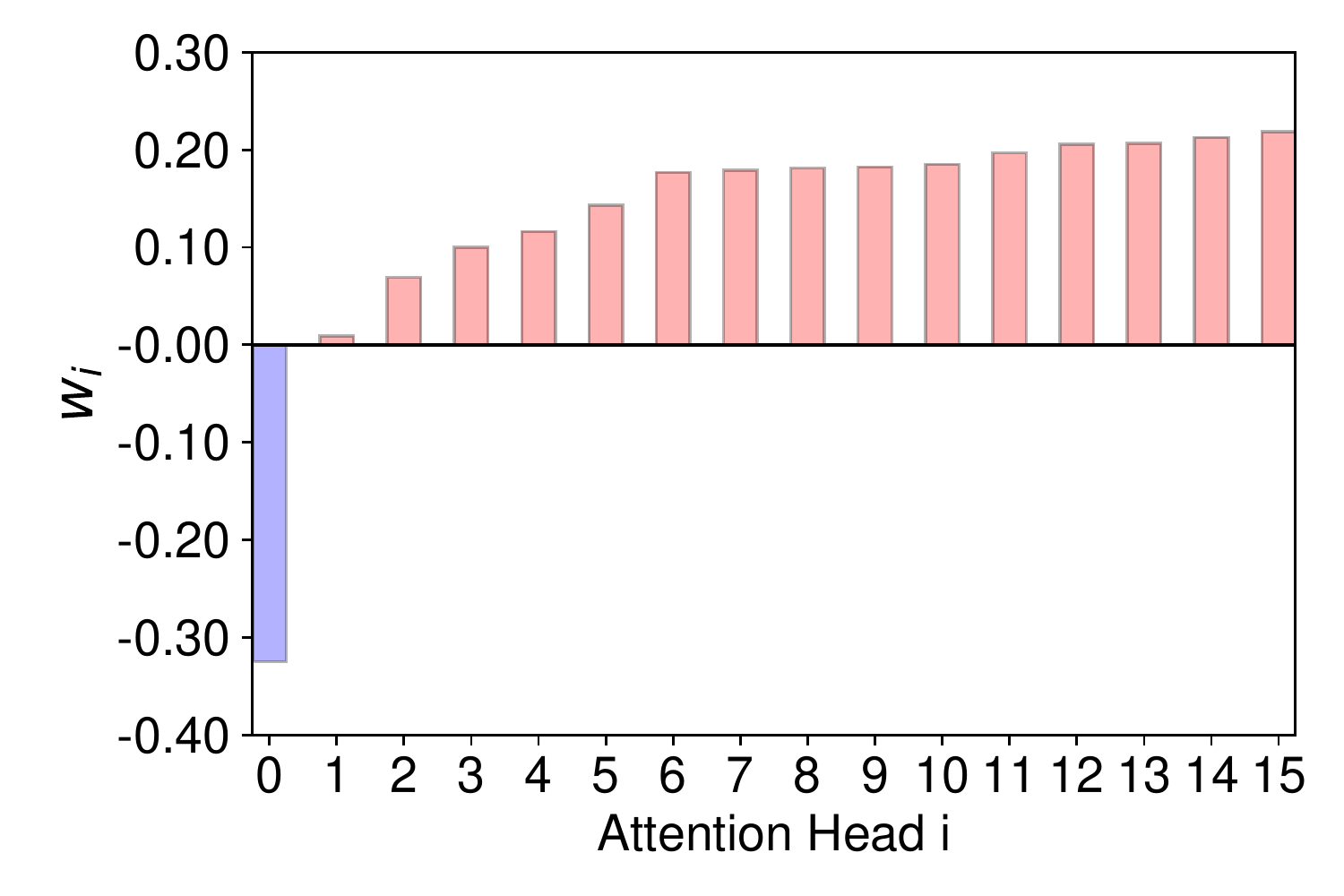} \label{fig.newshead2}
	}
		\subfigure[Word-level $v_i$.]{
	\includegraphics[width=0.22\textwidth]{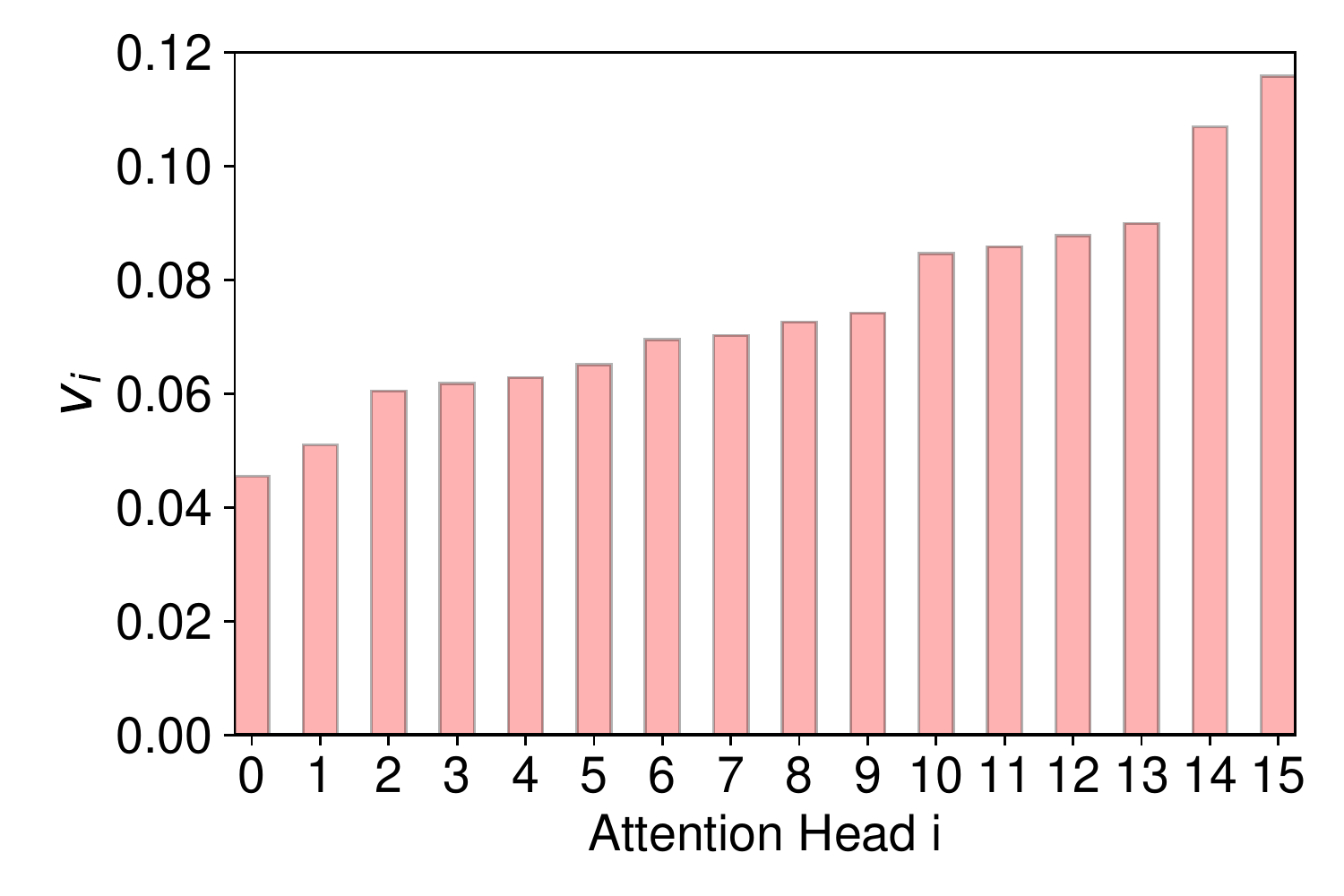} \label{fig.newshead3}
	}
		\subfigure[News-level $v_i$.]{
	\includegraphics[width=0.22\textwidth]{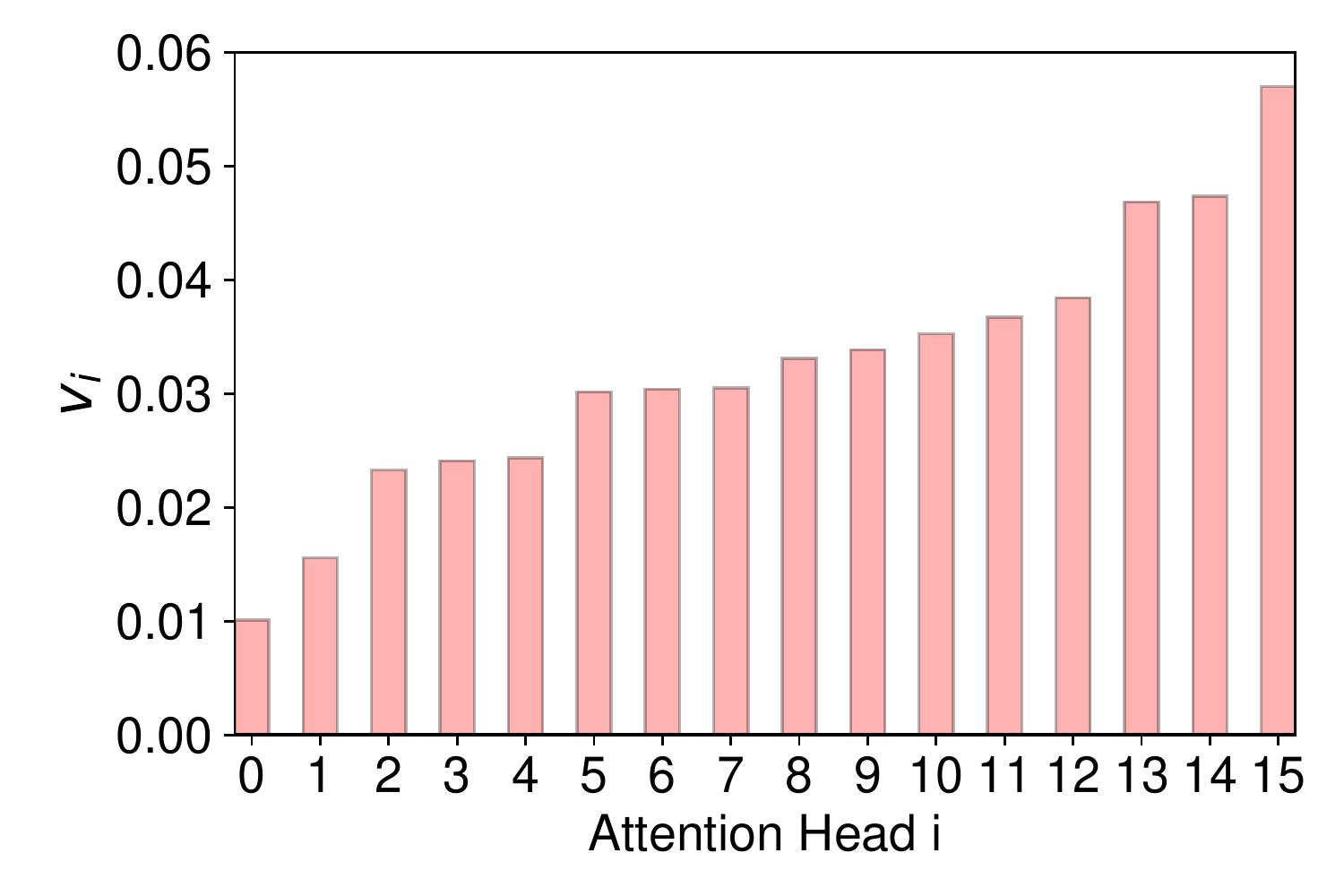} \label{fig.newshead4}
	}
\caption{The  distance weights learned by different attention heads on the \textit{MIND} dataset.}\label{fig.newshead}
\end{figure}

\begin{figure*}[!t]
	\centering 
	\subfigure[Vanilla Transformer.]{
	\includegraphics[width=0.98\textwidth]{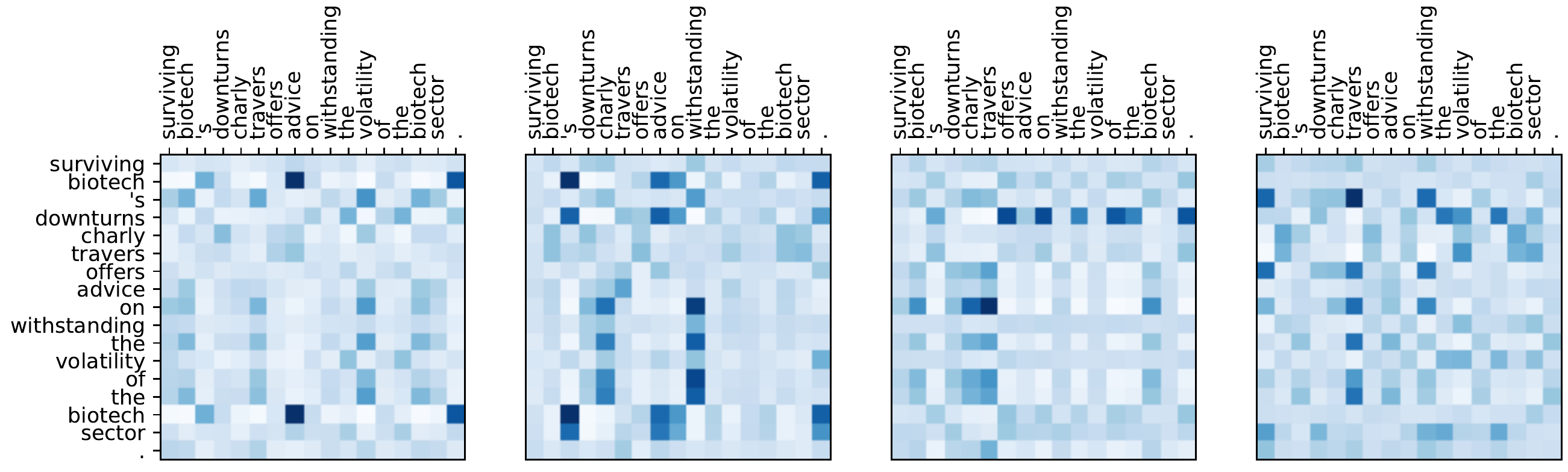} \label{fig.att1}
	}
		\subfigure[DA-Transformer. The first two heatmaps are produced by  heads with $w_i<0$ and others are produced by heads with $w_i>0$.]{
	\includegraphics[width=0.98\textwidth]{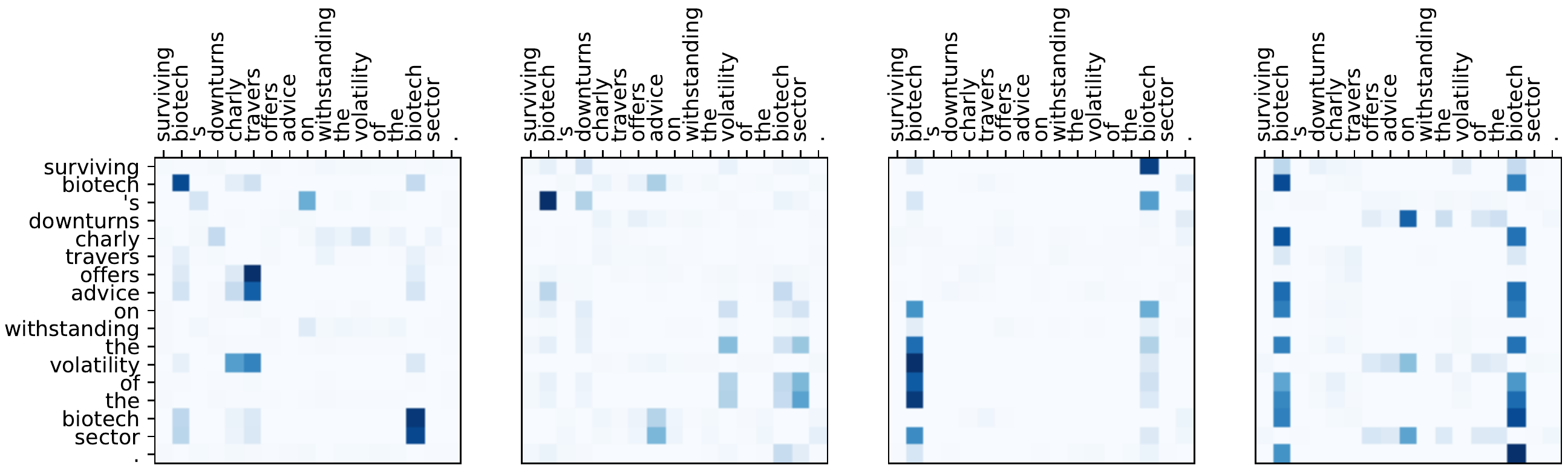} \label{fig.att2}
	}
\caption{The self-attention weights learned by the vanilla Transformer and our proposed DA-Transformer method. }\label{fig.heatmap}
\end{figure*}

\subsection{Influence of Different Mapping Functions}
Next, we study the influence of using different mapping functions $f(\cdot)$ for computing the re-scaled coefficients.
We compare the performance of our method w.r.t. several different $f(\cdot)$, including: (1) $f(x)=min(x,T)$ (clip), using a threshold $T$ to clip the weighted distance; (2) $f(x)=k_ix+b_i$ (linear), using a linear transformation to the weighted distance; (3) $f(x)=\mathrm{exp}(x)$ (exponent), using an exponent function to map the weighted distance; (4) $f(x)=\frac{1}{1+\mathrm{exp}(-x)}$ (sigmoid), using the sigmoid function to activate the weighted distance; and (5) $f(x;v_i)=\frac{1+\mathrm{exp}(v_i)}{1+\mathrm{exp}(v_i-x)}$, our learnable sigmoid function.
Due to space limitation, we only present the results on the \textit{AG}, \textit{Amazon} and \textit{MIND} datasets in Fig.~\ref{fig.positionfunc1}.
From these results, we find that clip is not optimal for mapping the weighted distance.
This is because it cannot keep the precise distance information beyond a certain range.
In addition, simply using the linear transformation is also insufficient.
This may be because our attention adjustment method requires $f(\cdot)$ to be positive, but linear transformation cannot guarantee.
Besides, we find that the sigmoid function and our proposed function are better than the exponential function.
This may be because long sequences will lead to the problem of exponent explosion, which is harmful to context modeling.
Moreover, our proposed learnable sigmoid function is better than the standard sigmoid function.
It shows that adjusting the activation function in a learnable way can better map the raw distances into re-scaled coefficients.

\subsection{Influence of Different Attention Adjusting Methods}

Then, we explore the influence of different methods for adjusting the raw attention weights.
We consider four different kinds of methods, including:  (1) adding the re-scaled coefficients to the attention weights normalized by softmax (late add); (2) multiplying the re-scaled coefficients with the attention weights normalized by softmax (late multiply); (3) adding the re-scaled coefficients to the raw attention weights before normalization (early add), which is widely used in existing methods like \textit{Transformer-XL}; (4) multiplying the re-scaled coefficients with the raw attention weights activated by ReLU, which is the method used in our approach (early multiply).
The results on the \textit{AG}, \textit{Amazon} and \textit{MIND} datasets are shown in Fig.~\ref{fig.positionfunc2}.
According to these results, we find that early adjustment is better than late adjustment.
This may be because the late adjustment methods will change the total amount of attention, which may not be optimal.
In addition, we find that multiplying is better than adding for both early and late adjustment.
This may be because adding large re-scaled coefficients may over-amplify some attention weights.
For example, if a raw attention weight is relatively small, it is not suitable to add large re-scaled coefficients to it because the corresponding contexts may not have close relations.
In contrast, multiplying the re-scaled coefficients will not over-amplify the low  attention weights.
Moreover, in our  early multiply method we further propose to use the ReLU function to introduce sparsity to make the Transformer more ``focused''.
Thus, our method is better than the existing early add method in adjusting the attention weights.

\subsection{Model Interpretation}

Finally, we interpret our proposed method by visualizing its key parameters and the attention weights.
we first visualize the parameters $w_i$ and $v_i$ in our method, which control the preferences of attention heads on long-term or short-term information and the shape of the learnable sigmoid function, respectively.
The visualization results on the \textit{AG} and \textit{MIND} datasets are respectively shown in Figs.~\ref{fig.aghead} and \ref{fig.newshead}.\footnote{We show the average results of 5 runs. The values of $w_i$ and $v_i$ in these figures are sorted and are not corresponding to the head orders.}
From Fig.~\ref{fig.aghead}, we find it is very interesting that half of the parameters $w_i$ are positive and the rest of them are negative.
It indicates that half of the attention heads mainly aim to capture local contexts, while the rest ones are responsible for modeling long-distance contexts.
It may be because both short-term and long-term contexts are useful for understanding news topics.
In addition, we find that most attention heads have negative $v_i$ while the rest are positive.
It shows that on the \textit{AG} dataset the intensity of attention adjustment is mild in most attention heads.
From Fig.~\ref{fig.newshead1}, we find long-term information is somewhat more important than local information in modeling news texts for news recommendation.
However, from Fig.~\ref{fig.newshead2} we find an interesting phenomenon that only one head has a strong negative $w_i$ while the values of $w_i$ in all the rest heads are positive.
It means that only one attention head tends to capture short-term user interests while all the other heads prefer to capture long-term user interests.
This is intuitive because users usually tend not to intensively click very similar news and their long-term interests may have more decisive influence on their news clicks.
In addition, we find it is interesting that on \textit{MIND} all values of $v_i$ are positive.
It may indicate that distance information has a strong impact on the attention weights.
These visualization results show that DA-Transformer can flexibly adjust its preference on short-term or long-term information and the intensity of attention adjustment by learning different values of $w_i$ and $v_i$ according to the task characteristics.\footnote{We do not observe significant correlations between the sequence length and the signs of $w_i$. This may indicate that the values of $w_i$ depend more on the task characteristics rather than text lengths.}

We then visualize the attention weights produced by the vanilla Transformer and the distance-aware attention weights in our DA-Transformer method.
The attention weights of a sentence in the \textit{AG} dataset computed by four different attention heads are respectively shown in Figs.~\ref{fig.att1} and \ref{fig.att2}.
From Fig.~\ref{fig.att1}, we find it is difficult to interpret the self-attention weights because they are too ``soft''. 
In addition, it is difficult for us to understand the differences between the information captured by different attention heads.
Different from the vanilla Transformer, from Fig.~\ref{fig.att2} we find that the attention weights obtained by our method are more sparse, indicating that the attention mechanism in our method is more focused.
In addition, it is easier for us to interpret the results by observing the attention heatmap.
For example, the first two heatmaps in Fig.~\ref{fig.att2} are produced by the two attention heads with preferences on short-term contexts.
We can see that they mainly capture the relations among local contexts, such as the relations between ``biotech'' and ``sector''.
Differently, in the latter two heatmaps obtained by the two attention heads that prefer long-term contexts, we can observe that the model tends to capture the relations between a word (e.g., ``biotech'') with the global contexts.
These results show that different attention heads in our method are responsible for capturing different kinds of information, and their differences can be directly observed from the self-attention weights.
Thus, our method can be better interpreted than vanilla Transformers.

\section{Conclusion}\label{sec:Conclusion}

In this paper, we propose a distance-aware Transformer, which can leverage the real distance between contexts to adjust the self-attention weights for better context modeling.
We propose to first use different learnable parameters in different attention heads to weight the real relative distance between tokens.
Then, we propose a learnable sigmoid function to map the weighted distances into re-scaled coefficients with proper ranges.
They are further multiplied with the raw attention weights  that are activated by the ReLU function to keep non-negativity and produce sharper attention.
Extensive experiments on five benchmark datasets show that our approach can effectively improve the performance of Transformer by introducing real distance information to facilitate context modeling.


\section*{Acknowledgments}
This work was supported by the National Natural Science Foundation of China under Grant numbers U1936208 and U1936216.

\bibliography{main}
\bibliographystyle{acl_natbib}

\end{document}